\documentclass[preprint,12pt]{elsarticle}

\usepackage{amssymb}
\usepackage{amsmath}

\journal{Information Fusion}
\usepackage{times}
\usepackage{hyperref}
\usepackage{url}
\usepackage[utf8]{inputenc}
\usepackage[small]{caption}
\usepackage{graphicx}
\usepackage{amsmath}
\usepackage{amsthm}
\usepackage{booktabs}
\usepackage{algorithm}
\usepackage{algorithmic}
\usepackage{xcolor}
\usepackage[switch]{lineno}
\usepackage{pax}
\usepackage{multirow}
\usepackage{rotating}
\usepackage{amssymb} 
\usepackage{tcolorbox}
\usepackage{soul}
\usepackage{adjustbox}
\usepackage{tabularx}
\usepackage{pifont}    
\usepackage{makecell}

\begin{document}
\begin{frontmatter}
\biboptions{sort&compress}

\title{Large Multimodal Models for Low-Resource Languages: A Survey}

\author[unibuc]{Marian Lupa\c{s}cu}
\author[unibuc]{Ana-Cristina Rogoz}
\author[unibuc]{Mihai Sorin Stupariu}
\author[unibuc]{Radu Tudor Ionescu}

\affiliation[unibuc]{organization={Department of Computer Science, University of Bucharest},
            country={Romania}}           


\begin{abstract}
In this survey, we systematically analyze techniques used to adapt large multimodal models (LMMs) for low-resource (LR) languages, examining approaches ranging from visual enhancement and data creation to cross-modal transfer and fusion strategies. Through a comprehensive analysis of 117 studies across 96 LR languages, we identify key patterns in how researchers tackle the challenges of limited data and computational resources. We categorize works into resource-oriented and method-oriented contributions, further dividing contributions into relevant sub-categories. We compare method-oriented contributions in terms of performance and efficiency, discussing benefits and limitations of representative studies. We find that visual information often serves as a crucial bridge for improving model performance in LR settings, though significant challenges remain in areas such as hallucination mitigation and computational efficiency. In summary, we provide researchers with a clear understanding of current approaches and remaining challenges in making LMMs more accessible to speakers of LR (understudied) languages. We complement our survey with an open-source repository available at: \url{https://github.com/marianlupascu/LMM4LRL-Survey}.
\end{abstract}

\end{frontmatter}

\section{Introduction}









Recent advancements in large multimodal models (LMMs) showcased remarkable capabilities in processing and understanding diverse types of data, including text, images, audio and video. Models like GPT-4V, KOSMOS-1 \citep{NEURIPS2023_e425b75b} and PaLM-E \citep{driess2023palm} achieved impressive performance levels across various multimodal tasks through their ability to simultaneously process and reason about multiple modalities. However, these developments have primarily focused on high-resource languages, particularly English, leaving a significant gap in supporting the world's many low-resource languages.
The distinction between high-resource (HR) and low-resource (LR) languages is primarily determined by the availability of digital resources and training data. High-resource languages, such as English, Mandarin, and Spanish, benefit from extensive digital corpora, parallel texts, and annotated datasets. In contrast, low-resource or understudied languages, which constitute the majority of the world's languages, lack sufficient digital resources, standardized datasets, and computational tools. This disparity is particularly pronounced in multimodal contexts, where the scarcity of paired data across modalities (e.g.~image-text pairs, audio-text alignments) poses additional challenges.

A recent analysis~\cite{dotan2024invisible} identified $27\%$ of languages as ``Invisible Giants'', i.e.~demographically robust yet digitally absent, highlighting that resource scarcity is institutionally constructed rather than inherent. This distinction has practical implications, e.g.~LMM development that treats data scarcity merely as technical risks can perpetuate the structural inequalities it ostensibly addresses. We therefore situate our analysis within the UNESCO International Decade of Indigenous Languages (2022-2032) and the CARE principles for Indigenous data governance~\cite{carroll2020care}, which emphasize community authority over linguistic data. Indeed, the very terminology ``low-resource'' has been critiqued as colonial and Eurocentric, obscuring the political decisions that produced linguistic marginalization~\cite{bird2022decolonising}.

The motivation for developing multimodal capabilities for LR languages is compelling. First, multimodal processing better reflects how humans naturally communicate and understand information through multiple sensory channels. Second, visual and audio cues can provide crucial contextual information that helps to overcome the limitations of scarce textual data. Third, many LR languages are primarily spoken rather than written, making multimodal approaches particularly relevant for their digital preservation and processing.
However, developing multimodal systems for LR languages faces several significant challenges, including: (1) the scarcity of high-quality multimodal datasets in these languages, (2) the lack of standardized evaluation benchmarks, (3) the computational cost of training large-scale models with limited resources, and (4) the complexity of handling different writing systems, dialects, and cultural contexts. Moreover, the problem of catastrophic forgetting \citep{kirkpatrick2017overcoming} when adapting pre-trained models to new languages and the challenge of maintaining performance across different modalities pose significant technical hurdles.

\noindent\textbf{Literature selection process.}
We survey research articles from 2018 to 2025 that specifically study LMMs for LR languages. We begin our analysis with 2018 because one of the first large language models (LLMs), BERT \citep{devlin-etal-2019-bert}, was introduced that year, marking a significant turning point in the development of modern language modeling techniques. We focus on works that go beyond simple cross-lingual transfer or translation, examining techniques that leverage multiple modalities to improve model performance. 

\begin{figure}[!t]
     \centering
    \includegraphics[width=1.0\linewidth]{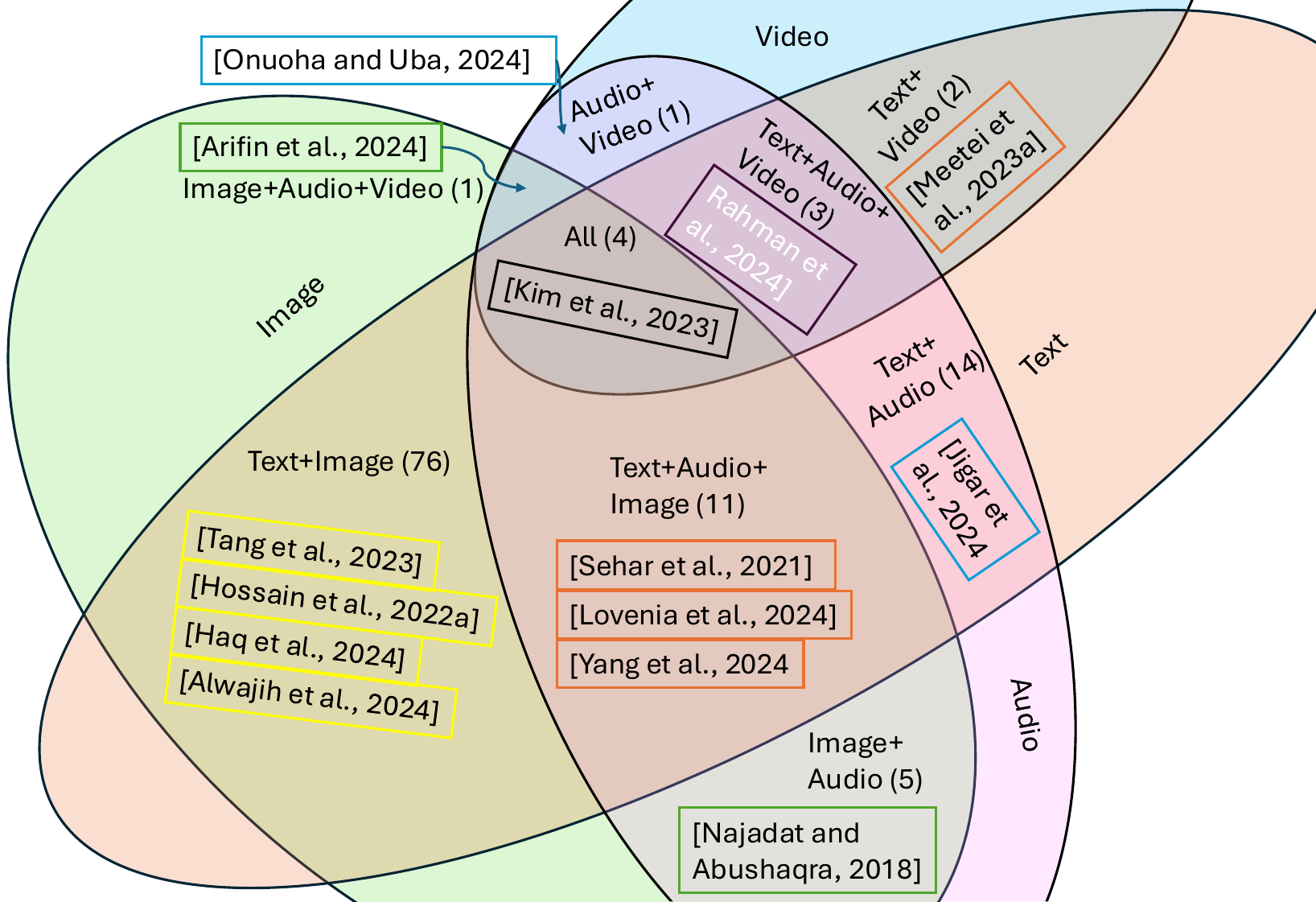}
    \vspace{-0.6cm}
    \caption{A Venn diagram with the distribution of papers across different modality combinations used by LMMs for low-resource languages. Text+image is the dominant modality pair, while more complex video-inclusive combinations are less common. A selection of representative papers is included for each  modality combination. References are clickable links to papers.}
    \label{fig:venn_chart}
\end{figure}


We queried a broad set of digital libraries to ensure representative coverage: ACM Digital Library, IEEE Xplore, ACL Anthology, arXiv, SpringerLink, ScienceDirect, and Google Scholar. During this search, we specifically targeted venues known for frequent LR or multimodal contributions (e.g.~ACL, EMNLP, NAACL, COLING, LREC, EACL, WMT, CVPR/ICCV/ECCV workshops, and INTERSPEECH) to ensure that relevant conference and workshop publications are captured.
For each of these digital libraries, we formulated several keyword combinations capturing (i) multimodality, (ii) low-resource aspects, and (iii) language or task types. To further improve coverage, we performed a backward and forward search to identify additional relevant work from the reference lists of included papers, and we used citation links to identify more recent follow-up studies.

\begin{figure}[t!]
     \centering
    \includegraphics[width=1.0\linewidth]{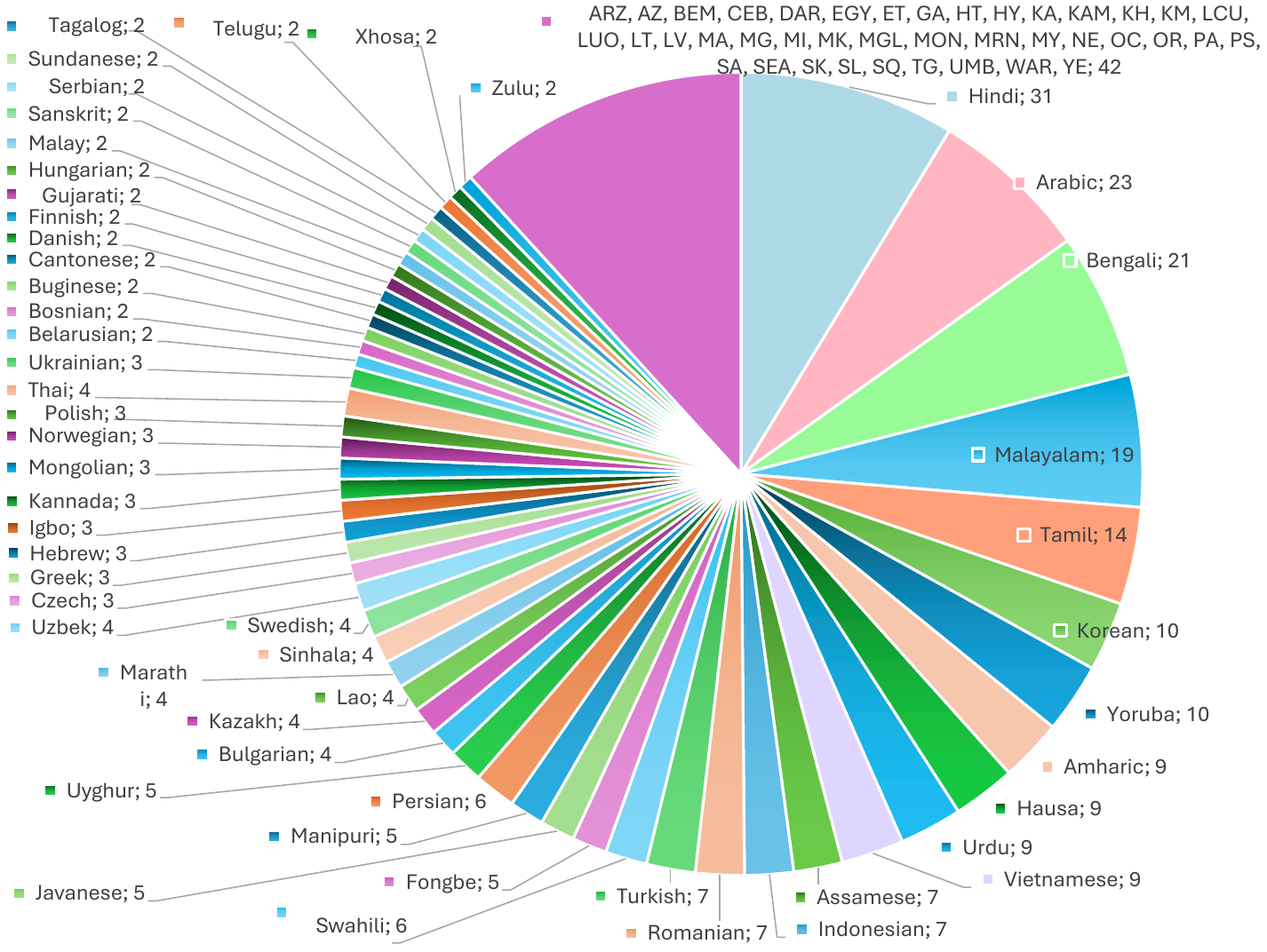}
    \vspace{-0.5cm}
    \caption{Distribution of papers across 96 low-resource languages, representing 117 papers. Hindi leads with 31 studies, followed by Arabic (23), Bengali (21),  Malayalam (19), Tamil (14), Korean and Yoruba (with 10 papers each). 
    The remaining languages have less than 10 papers each. Languages with only one paper (42 languages) are listed using ISO 639-1 codes.
    The data highlights the disparity in research focus among LR languages, with a few languages receiving more focus, while many others remain understudied in the context of multimodal learning. Some papers simultaneously address multiple languages, contributing to the individual language counts. HR languages such as English, Chinese, Mandarin and Spanish are excluded from this chart.}
    \label{fig:pie_chart}
    \vspace{-0.4cm}
\end{figure}

Finally, we merged all retrieved records and applied a manual two-stage screening process. We began by reviewing titles and abstracts to remove clearly irrelevant work (e.g.~single modality studies or studies exclusively targeting high-resource languages such as English, Mandarin, or Spanish). We then examined the main contributions of the remaining papers to assess whether they were a suitable fit for our survey. Ultimately, a study was included in our survey if it matched all of the following criteria: (a) is multimodal (at least two input modalities), (b) focuses on low-resource languages (at least one of the targeted languages was an underrepresented language), and (c) proposes, adapts or evaluates a multimodal model.

\noindent\textbf{Research focus distribution across LR languages.}
Our survey reveals several interesting patterns in how researchers approached multimodal learning for LR languages. As shown in Figure \ref{fig:venn_chart}, text-image combinations dominate the research landscape, appearing in 76 papers (65\% of surveyed works), while more complex combinations incorporating audio and video remain less explored. In addition, the distribution of research focus across languages is notably uneven, as illustrated in Figure \ref{fig:pie_chart}, with Hindi (31 papers), Arabic (23) and Bengali (21 papers) receiving significant focus, whereas 42 other languages are each represented by a single study. On the one hand, this striking disparity highlights the need for a broader coverage of understudied languages in multimodal research. On the other hand, it also warrants critical examination of the factors influencing the distribution of research studies across languages.

\begin{table}[!t]
\centering
\caption{Factors explaining research disparity across low-resource languages in multimodal NLP. 
We categorize languages from our survey by paper count and analyze contributing factors.}
\label{tab:disparity_factors}
\begin{adjustbox}{max width=\textwidth}
\small
\begin{tabular}{@{}p{3.3cm}p{4.5cm}p{4.5cm}p{4.5cm}@{}}
\toprule
\textbf{Factor} & \textbf{High Coverage} (10+ papers) & \textbf{Medium Coverage} (2-9 papers) & \textbf{Low Coverage} (1 paper) \\
\midrule
Institutional capacity & 
Strong local NLP communities (India, Middle East) & 
Emerging research groups & 
Minimal local infrastructure\\
\addlinespace
Speaker population & 
Variable (38M--600M) & 
Variable (2M--200M) & 
Often $<$1M, but not always \\
\addlinespace
Existing resources & 
Multiple benchmarks, pre-trained models & 
Some datasets available & 
Little to no digital presence \\
\addlinespace
Script accessibility & 
Shared with HR languages or well-supported & 
Moderate tool support & 
Often unique / unsupported scripts \\
\addlinespace
Geographic location & 
Regions with NLP venues (Asia, Middle East) & 
Mixed & 
Often Sub-Saharan Africa, Oceania \\
\addlinespace
Geopolitical interest & 
Strategic priority (Arabic post-9/11, Hindi for US-India relations, Korean for East Asia security) & 
Emerging strategic relevance (Ukrainian post-2022, Turkish for NATO relations) & 
No perceived strategic value; excluded from defense / intelligence funding \\
\addlinespace
Example languages & 
Hindi, Arabic, Bengali, Malayalam & 
Swahili, Romanian, Turkish, Yoruba & 
Luo, Xhosa, Occitan, Maori \\
\bottomrule
\end{tabular}
\end{adjustbox}
\end{table}

We identify six interacting factors that explain this disparity (see Table~\ref{tab:disparity_factors}). 
First, \textbf{institutional research capacity} plays a dominant role: Rungta et al.~\cite{rungta2022geographic} demonstrated that NLP publications are heavily concentrated in North America, Western Europe, and China, 
with minimal representation from Africa and South America. Languages spoken in regions with established 
NLP research communities (e.g.~Hindi in India, Arabic in the Middle East) benefit from existing 
infrastructure, funding, and researcher networks. Second, \textbf{speaker population} shows surprising variability: 
while one might expect larger speaker populations to attract more research, this correlation is weak. For instance, Swahili 
has approximately 200 million speakers, yet remains underrepresented compared with Malayalam (38 million speakers, 
19 papers). Third, \textbf{digital resource availability} creates 
a self-reinforcing cycle: languages with existing datasets attract more research, which produces more 
datasets~\cite{joshi2020state}. Ranathunga et al.~\cite{ranathunga2022languages} showed that even within 
the same resource class~\cite{joshi2020state}, languages from higher-GDP regions receive disproportionately more 
research attention. Fourth, \textbf{script and typological proximity} to high-resource languages facilitates transfer learning research, e.g.~Hindi benefits from shared Devanagari script resources, while languages with 
unique scripts (e.g.~Ge'ez for Amharic) face additional barriers. Fifth, \textbf{geographic location} 
determines access to NLP venues and research networks: languages spoken in regions hosting major conferences 
(Asia, Middle East, Europe) receive more attention than those in Sub-Saharan Africa or Oceania. Sixth, 
\textbf{geopolitical interest} drives strategic investment: Arabic NLP surged post-9/11, Ukrainian gained 
attention after 2022, and US-China AI competition benefits research on Mandarin Chinese, but not on minority languages within China.

These factors have critical implications for researchers working on truly underrepresented languages. 
The 42 languages with single studies in our survey face a ``cold start'' problem: without existing 
benchmarks, baselines, or community momentum, new contributions are harder to contextualize and 
evaluate~\cite{caines2019geographic}. We observe that 88.4\% of the world's languages (Class 0 defined by Joshi et al.~\cite{joshi2020state}) have no representation in standard NLP resources whatsoever~\cite{joshi2020state}. For 
researchers targeting these languages, we recommend: (1) prioritizing dataset creation with community 
involvement over model development, (2) leveraging typologically similar languages for transfer rather 
than defaulting to English, and (3) publishing in venues with explicit low-resource tracks 
(e.g.~AfricaNLP, AmericasNLP, etc.) to build critical mass within language-specific research communities.

Research investment in specific languages is strongly 
influenced by geopolitical events and national security priorities. The clearest documented case is 
Arabic NLP, which experienced a dramatic surge in funding following September 11, 2001. Darwish et 
al.~\cite{habash2022panoramic} documented the fact that ``Arabic NLP gained increasing importance in the Western 
world especially after September 11. The USA funded large projects for companies and research centers 
to develop NLP tools for Arabic and its dialects''. This investment wave (2001-2010) produced 
fundamental resources for machine translation, speech recognition, named entity recognition, and 
information extraction, that continue to underpin Arabic multimodal research today.

A similar pattern is emerging for Ukrainian. Prior to February 2022, Ukrainian was a moderately-resourced Slavic language, receiving limited attention in NLP research. The Russian invasion triggered rapid mobilization: 
the CLARIN Knowledge Centre for Ukrainian NLP (UkrNLP-Corpora) was established in 2023~\cite{clarin2023ukrainian}, the Ukrainian Natural Language Processing Workshop (UNLP) expanded to four editions by 2025, and 
researchers developed numerous datasets for disinformation detection, sentiment analysis, and propaganda identification on Ukrainian social media~\cite{unlp2025proceedings}. This research is explicitly framed around information warfare: detecting ``manipulative narratives'' and ``rhetorical manipulation techniques used to influence Ukrainian Telegram users'' \cite{akhynko2025hidden}. The geopolitical urgency has attracted Western funding and research attention that Ukrainian might not otherwise have received.

The US-China technology competition further illustrates how strategic rivalries shape NLP investment 
trajectories. China invested \$125 billion in AI in 2025, representing 38\% of 
global AI investment, with NLP receiving 11\% of this allocation~\cite{secondtalent2025chinese}. 
Chinese companies, including Baidu, Alibaba, and Tencent, are developing competitive large language models 
(Qwen, Yi, DeepSeek) trained on massive Chinese corpora, partly driven by the US export controls on 
advanced semiconductors~\cite{science2025deepseek}. This competition benefits Mandarin Chinese language resources, 
but does not extend to minority languages within China (Tibetan, Uyghur, Mongolian), which remain severely 
underrepresented despite large speaker populations. This indicates that geopolitical attention flows to 
languages of strategic interest to major powers, not necessarily to the most linguistically marginalized 
communities.

The patterns identified above reveal a troubling dynamic for truly underrepresented languages: research investment 
follows geopolitical salience rather than linguistic need. Languages become ``high-resource'' when 
powerful states perceive strategic value in processing them for intelligence gathering, countering 
disinformation, or economic competition. The 42 languages in our survey with only a single study 
lack this geopolitical visibility. For researchers working on such languages, this suggests that 
framing research around emerging strategic concerns (e.g.~climate migration, regional stability, 
pandemic communication) may attract funding that purely linguistic motivations cannot.

\begin{table*}[t!]
\centering
\caption{Comparison of our survey with related work on LLMs and LMMs across different focuses, modalities, languages, techniques, and additional coverage.}
\label{tab:survey_comparison}
\resizebox{\textwidth}{!}{%
\setlength\tabcolsep{0.1cm}
\begin{tabular}{p{3.2cm}p{4.5cm}cp{2.5cm}cccccp{6.5cm}}
\toprule
\multirow{7}{*}{\textbf{Survey}} & \multirow{7}{*}{\textbf{Focus}} & \multirow{2}{*}{\textbf{\rotatebox{90}{Multimodality$\;\;\;$}}} & \multirow{7}{*}{\textbf{Languages}} & \multicolumn{5}{c}{\textbf{Techniques}} & \multirow{7}{*}{\textbf{Additional Coverage}} \\
\cmidrule(lr){5-9}
& & & & \textbf{\rotatebox{90}{Data Creation}} & \textbf{\rotatebox{90}{Fusion}} & \textbf{\rotatebox{90}{Visual Enh.}} & \textbf{\rotatebox{90}{Transfer}} & \textbf{\rotatebox{90}{Adaptation}} & \\
\midrule
Gu et al.~\cite{gu2024survey}           & LLM-as-a-judge               &                         & High-resource &                          &                         &                         &                         &                         & Evaluation, reliability, applications \\
Joshi et al.~\cite{joshi2020state}         & Language resources           &                         & Low-resource  & \checkmark               &                         &                         &                         &                         & Digital divide \\
Paullada et al.~\cite{Paullada2020DataAI}     & Dataset development          &                         & Low-resource  & \checkmark               &                         &                         &                         &                         & Data challenges \\
Ruder et al.~\cite{Ruder2021XTREMERTM}     & Cross-lingual NLP            &                         & Low-resource  &                          &                         &                         & \checkmark              &                         & Benchmarking \\
Zhao et al.~\cite{Zhao2023ASO}            & LLMs                         &                         & Both  & \checkmark               &                         &                         & \checkmark              & \checkmark              & Pre-training, adaptation, utilization \\
Zhu et al.~\cite{Zhu2023ExtrapolatingLL} & Multilingual LLM             &                         & Both          & \checkmark               &                         &                         & \checkmark              &                         & Cross-lingual transfer \\
Gan et al.~\cite{Gan2022VisionLanguagePB} & Vision-language             & \checkmark              & High-resource & \checkmark               & \checkmark              & \checkmark              &                         &                         & Pre-training objectives \\
Wang et al.~\cite{2023-Wang-survey}       & Pre-training                 & \checkmark              & High-resource & \checkmark               &                         &                         &                         & \checkmark              & Model architectures \\
Li et al.~\cite{Art_Large_Vision}       & Large VLMs & \checkmark              & High-resource &                          &                         &                         &                         &                         & Benchmark evaluations, challenges \\
Yin et al.~\cite{2023-yin}               & LMM architectures            & \checkmark              & High-resource & \checkmark               & \checkmark              &                         &                         & \checkmark              & Training strategies \\
Xie et al.~\cite{xie2024large}           & Large multimodal agents      & \checkmark              & High-resource &                          &                         &                         &                         &                         & Agentic AI, evaluation methods \\
Xu et al.~\cite{xu2024survey}           & Resource-efficient models    & \checkmark              & Both          &                          &                         &                         &                         & \checkmark              & Efficient algorithms, system designs \\
Alam et al.~\cite{2024-alam}              & LLMs for LR contexts & \checkmark            & Low-resource  &                          &                         &                         & \checkmark              & \checkmark              & Capabilities, prompting, evaluation \\

Mu et al.~\cite{DBLP:journals/corr/abs-2503-07137} & Mixture-of-Experts & \checkmark & Both &                          &                         &                         &                         & \checkmark              & Algorithms, theory, applications \\

\midrule
\textbf{Our Survey}              & \textbf{LMMs for LR languages} & \textbf{\checkmark}   & \textbf{Low-resource} & \textbf{\checkmark}     & \textbf{\checkmark}     & \textbf{\checkmark}     & \textbf{\checkmark}     & \textbf{\checkmark}     & \textbf{Systematic taxonomy, 96 languages, 117 studies} \\
\bottomrule
\end{tabular}%
}
\end{table*}

\begin{figure*}[t!]
    \centering
    \includegraphics[width=1.0\textwidth]{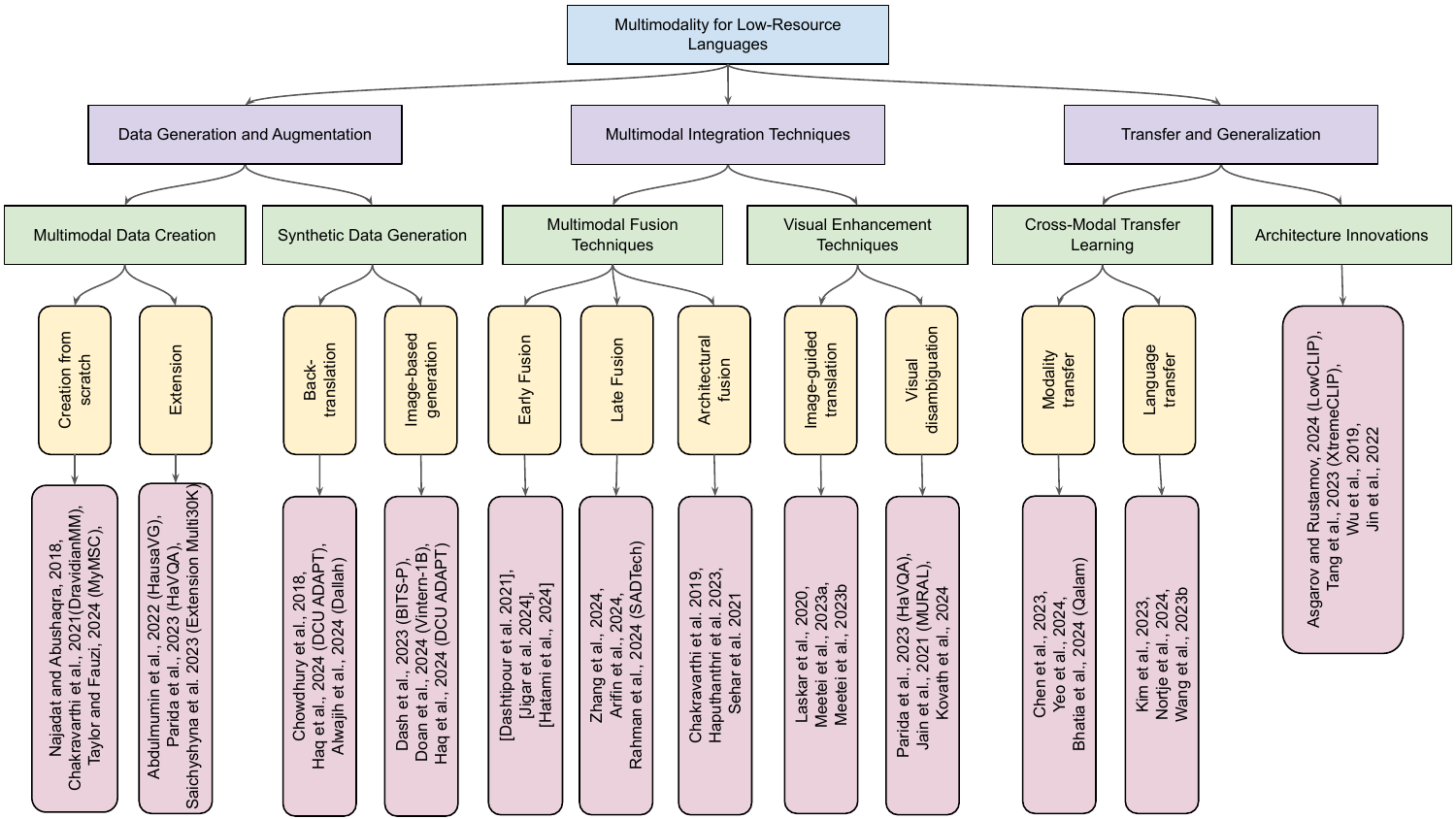}
    \vspace{-0.75cm}
    \caption{High-level taxonomy of LMMs for low-resource languages. We depict six main categories (inside boxes with green background), which are further divided into subcategories, exemplified via a few representative studies. References are clickable links to papers.}
    \label{fig:Taxonomy}
\end{figure*}

\noindent\textbf{Relation to other surveys and academic contributions.}
Some recent surveys have explored various aspects of multimodal language models. Zhao et al.~\cite{Zhao2023ASO} provided a comprehensive overview of LMM architectures, training strategies, and applications, while Wang et al.~\cite{2023-Wang-survey} focused on pre-training techniques and model architectures. Additional surveys have examined related areas, including LLMs \citep{gu2024survey,joshi2020state,Paullada2020DataAI,Ruder2021XTREMERTM,Gan2022VisionLanguagePB,Art_Large_Vision,xie2024large,xu2024survey,DBLP:journals/corr/abs-2503-07137}, but they did not specifically address the unique challenges and solutions for LR languages in multimodal contexts. Alam et al.~\cite{2024-alam} explored LLMs for low-resource languages in multilingual, multimodal and dialectal settings, but they focused primarily on the capabilities of LLMs rather than presenting a comprehensive survey of techniques. To the best of our knowledge, our survey is the first to focus on multimodal learning for understudied languages.

As shown in Table~\ref{tab:survey_comparison}, our work differs from previous surveys by specifically focusing on the intersection of multimodality and low-resource languages, while addressing all major techniques relevant to this domain. While several prior surveys have separately explored multimodality or low-resource languages, none has comprehensively examined both aspects across such a diverse range of languages and approaches.

In summary, our contribution is fourfold:
\begin{itemize}
\item \vspace{-0.2cm} We provide the first comprehensive analysis of LMMs specifically focused on LR languages, examining 117 studies across 96 languages.
\item \vspace{-0.2cm} We develop a novel taxonomy (see Figure \ref{fig:Taxonomy}) that categorizes existing approaches into six main categories: multimodal data creation, synthetic data generation, multimodal fusion techniques, visual enhancement techniques, cross-modal transfer learning and architectural innovations.
\item \vspace{-0.2cm} We systematically organize the literature to enable a clear understanding of current approaches and remaining challenges in making LMMs more accessible to speakers of LR languages.
\item \vspace{-0.2cm} We provide an open-source repository that includes implementation details, datasets, and benchmarks to facilitate future research in this emerging field.
\end{itemize}

\noindent\textbf{Organization.}
The remainder of this survey is organized as follows. In Section {\ref{sec_taxo}}, we present an overview of the constructed taxonomy and discuss research trends between 2018 and 2025. Sections {\ref{sec_data_creation}} and {\ref{sec_data_gen}} are dedicated to resource-oriented contributions. In Section {\ref{sec_data_creation}}, we identify and categorize common approaches for dataset creation for LR languages. In Section {\ref{sec_data_gen}}, we discuss automated data generation techniques. Sections {\ref{sec_fusion}}, {\ref{sec_visual}}, {\ref{sec_cross}} and {\ref{sec_arch}} are dedicated to method-oriented contributions. In Section {\ref{sec_fusion}}, we categorize and compare strategies used to fuse multiple modalities. In Section {\ref{sec_visual}}, we discuss techniques used to enhance machine translation by using visual information. In Section {\ref{sec_cross}}, we present methods that perform cross-modal transfer learning. In Section {\ref{sec_arch}}, we analyze architectural contributions. In Section {\ref{sec_eval}}, we discuss current evaluation challenges and propose several ways to address the identified challenges. In Section {\ref{sec_conclusion}}, we draw our conclusions, point out current research gaps, and propose ways to mitigate them in future work.

\section{Taxonomy}
\label{sec_taxo}


To organize the diverse approaches in the rapidly evolving field of LMMs for low-resource languages, we develop a comprehensive taxonomy through a systematic analysis of the 117 papers in our survey. Our methodology involves initial coding of each paper's primary contributions and techniques, iterative refinement through thematic analysis to identify recurring patterns, and hierarchical organization of approaches based on their functional relationships and chronological development in the field.
Our analysis reveals that researchers addressing the challenges of LMMs for low-resource languages typically follow a progression from resource development to architectural refinement. This progression is reflected in our taxonomy, which organizes approaches into six main categories that represent both the current state of the field and the primary research strategies for addressing challenges in the context of underrepresented languages. 

In Figure \ref{fig:Taxonomy}, we systematically organize LMMs for LR languages into six main categories. The first two categories focus on constructing high-quality resources. While the first category discusses multimodal data creation either from scratch or via extending existing datasets, the second approach centers on synthetic data generation, which automatically expands available resources via back-translation and image-based generation. Building upon this work, we present several multimodal fusion techniques and provide various strategies for effectively combining this information, ranging from early and late fusion to more complex hybrid approaches. In the fourth category, we illustrate visual enhancement techniques that harness visual information through image-guided translation and visual disambiguation methods, highlighting their importance for improving translation quality and resolving ambiguities. Expanding from the single-modality solutions, the next category focuses on cross-modal transfer learning approaches that can facilitate knowledge sharing based on both modality transfer and language transfer. Finally, our last category comprises architectural innovations specifically tailored for multimodal tasks in the context of LR languages. 

It is important to note that several studies naturally span multiple areas. For instance, some studies that fuse visual and textual features could also be viewed as cross-modal transfer when they leverage pre-trained vision-language models. Similarly, papers that introduce new datasets may incorporate synthetic augmentation, and architecture-focused contributions may sometimes rely on transfer-learning mechanisms. In such cases, we assign each study to the category that reflects its primary technical contribution. This principle helps maintain clear boundaries, while acknowledging the natural overlap across multimodal methods for low-resource languages.

\begin{figure}[t]
    \centering
    \includegraphics[width=0.98\linewidth]{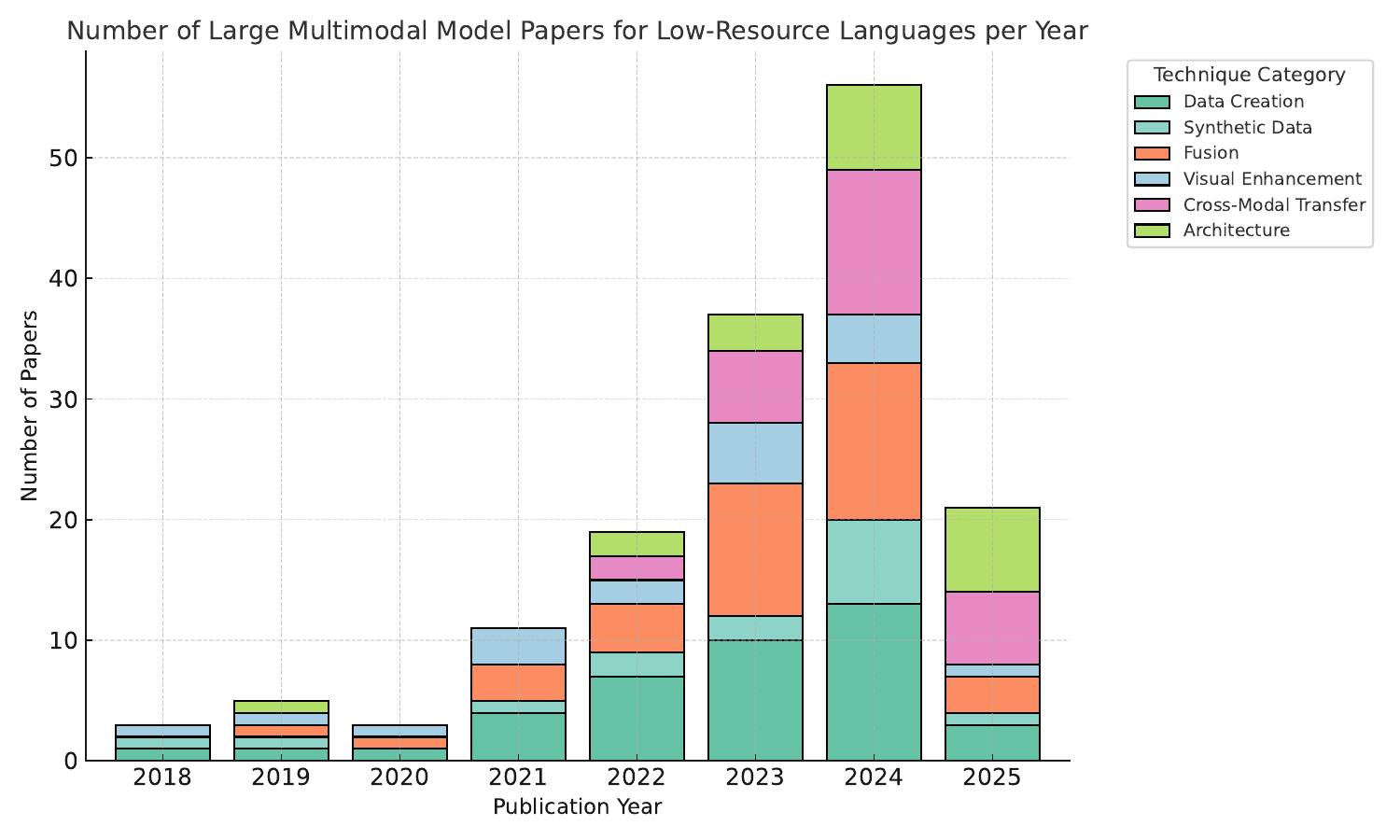}
    \caption{Number of LMM papers for LR languages published per year (2018-2025), categorized by technique: Multimodal Data Creation, Synthetic Data Generation, Multimodal Fusion Techniques, Visual Enhancement Techniques, Cross-Modal Transfer Learning, and Architectural Innovations. Best viewed in color.}
    \label{fig:lmm_taxonomy_trends}
\end{figure}

Furthermore, to understand how these categories have evolved over time, we analyze the publication trends from 2018 to 2025. Figure~\ref{fig:lmm_taxonomy_trends} shows the number of papers per year in each category, illustrating how early work focused primarily on data creation and synthetic augmentation, while recent years saw an increase in fusion strategies, cross-modal transfer, and architectural innovations, reflecting the shift toward foundation-model-based approaches.

Our taxonomic structure not only organizes existing research, but also highlights the inter-dependencies between different approaches and reveals gaps in current research, particularly in the exploration of complex multimodal combinations involving video and speech for low-resource languages. We structure the remainder of this article according to our novel taxonomy shown in Figure \ref{fig:Taxonomy}.

\section{Multimodal Data Creation}
\label{sec_data_creation}

There are two main approaches to create multimodal datasets for LR languages. The first is based on multimodal dataset creation from scratch, while the second is based on using an existing resource as a starting point. We next discuss papers introducing novel datasets based on the two alternatives.


\noindent
\textbf{Dataset creation from scratch.}
Dataset creation from scratch has emerged as a crucial approach for enabling multimodal research in LR languages, particularly for sentiment analysis and specific language tasks. Multiple research teams have focused on creating specialized datasets through direct data collection and annotation, such as collecting Arabic videos with multimodal features for sentiment analysis \citep{2018-najadat}, building comprehensive Tamil and Malayalam video review datasets \citep{2021-Chakravarthi}, developing new corpora for languages such as Malay \citep{2024-taylor}, creating speech translation resources for Fongbe \citep{kponou-etal-2024-ffstc} and compiling Arabic multimodal sentiment collections \citep{2023-haouhat}. A significant trend has been the creation of meme-based datasets, with efforts focused on Bengali, through MemoSen and MUTE \citep{2022-hossain,2022-hossain-mute}, and Romanian, through RoMemes \citep{Pais2024RoMemesAM}, all incorporating multiple levels of annotation.

These dataset creation efforts have expanded beyond sentiment analysis to encompass other crucial applications, such as sign language recognition with ArabSign \citep{2022-hamzah}, and multi-purpose datasets like BIG-C for Bemba \citep{2023-sikasote}. Additionally, the creation of a Manipuri-English parallel corpus with accompanying audio recordings for speech-to-text translation \citep{2021-sanayai} provides an important resource for research in low-resource languages. More recently, Farsi et al.~\cite{farsi-etal-2025-persian} introduced a comprehensive suite of multimodal datasets for Persian, covering tasks such as VQA, OCR, visual abstraction reasoning, and cultural knowledge grounding. These projects typically involve careful quality control by using multiple annotators, standardized recording environments, and expert validation, demonstrating a shift toward building comprehensive resources specifically designed for LR languages, rather than relying on translation or transfer from high-resource languages.

\noindent
\textbf{Dataset extension.}
In addition to building data from scratch in the context of LR language understanding, there have been several efforts for leveraging existing datasets of rich-resource languages and building upon them. Sen et al.~\cite{2022-sen} introduced the Bengali Visual Genome (BVG) dataset, which extends the Visual Genome dataset \citep{krishna2017visual} with Bengali translations and annotations, enabling the development and evaluation of multimodal models for Bengali-English machine translation (MT) and image captioning. Similarly, Abdulmumin et al.~\cite{2022-abdulmumin} created the Hausa Visual Genome (HaVG) dataset by translating a subset of the Visual Genome dataset into Hausa, providing a valuable resource for English-to-Hausa multimodal MT. Building upon prior work and continuing the focus on the Hausa language, Parida et al.~\cite{2023-parida} introduced the Hausa Visual Question Answering (HaVQA) dataset, which adapts question-answer pairs from the Visual Genome dataset to the Hausa language through manual translation, creating the first visual question-answering (VQA) dataset for Hausa. Extending this trend to Indian languages, Parida et al.~\cite{parida-etal-2025-ovqa} introduced OVQA, a multimodal dataset for the Odia language, by translating over 6,000 question-answer pairs and associated captions from the Visual Genome dataset into Odia. Similarly, Anwar et al.~\cite{2023-anwar} introduced MuAViC, a multilingual audio-visual corpus providing 1,200 hours of audio-visual speech across 9 languages, establishing the first open benchmark for audio-visual speech-to-text translation.

Apart from the focus on African languages, Saichyshyna et al.~\cite{2023-saichyshyna} extended the Multi30K dataset \citep{elliott2016multi30k} to include Ukrainian translations and captions, facilitating integrated vision and language research in Ukrainian. More recently, Lovenia et al.~\cite{2024-lovenia} presented SEACrowd, a comprehensive multilingual and multimodal data hub and benchmark suite for Southeast Asian languages, which covers 13 tasks across three modalities (text, image, and audio) and 38 Southeast Asian indigenous languages, while Lent et al.~\cite{lent-etal-2024-creoleval} introduced CreoleVal, an extensive collection of benchmarks for 28 Creole languages, addressing the significant resource gap for these historically marginalized language varieties.

\section{Synthetic Data Generation}
\label{sec_data_gen}


An alternative approach to efficiently create multimodal datasets for LR languages relies on synthetic data generation. Unlike traditional dataset creation, which typically involves intensive manual data collection, human annotation, and domain-specific curation, synthetic data generation leverages existing resources and automated techniques to produce new multimodal content with minimal human input. This distinction is critical, as synthetic methods offer a scalable alternative for low-resource settings, where manual annotation is often costly or infeasible.

\noindent
\textbf{Back-translation.}
A common approach for synthetic data generation relies on the usage of back-translation, which has proven to be an effective technique to enhance the data for multilingual MT (MMT) in LR language pairs. This technique works by translating text from an HR language into an LR language, and then back again, helping to generate additional aligned examples without requiring human involvement. Dutta Chowdhury et al.~\cite{2018-dutta} demonstrated the effectiveness of this technique for training a neural MMT system in the context of LR language pairs by leveraging the Flickr30k dataset \citep{young2014image} and translating the source-language (English) captions to the target LR language (Hindi). Meetei et al.~\cite{meetei2021low} extended this approach for low-resource multimodal neural machine translation in the news domain for English-Hindi. In the WMT24 English-to-Low-Resource Multi-Modal Translation task, Haq et al.~\cite{2024-haq} showcased the effectiveness of back-translation for Hindi. Another use case of back-translation was shown by Alwajih et al.~\cite{2024-Alwajih-dallah}, who, starting from English-based image-text pairs, employed translation to Arabic, as well as back-translation. This was necessary for evaluating the quality of the translation, before passing the data to humans for Arabic dialect translation and training a dialect-aware LMM, named Dallah. However, the consistency of back-translated synthetic data has been a concern. 
To address this issue, Wang et al.~\cite{2023-wang-adapting} proposed a framework to improve the robustness of models when adapting grounded VQA models to LR languages, aiming to improve the performance without relying on machine-translated data. Wang et al.~\cite{wang2022cross} further explored this challenge by introducing noise-robust learning for cross-lingual cross-modal retrieval to handle translation noise in machine-translated sentences.

\noindent
\textbf{Image-based generation.}
Another mainstream approach for synthetic data generation uses images as a starting point. In the case of Indic language multimodal MT \citep{2023-dash}, synthetic images generated by diffusion models were deemed beneficial, their main goal being that of capturing the complexity of the target domain, and augmenting the existing image dataset. Similarly, Haq et al.~\cite{2024-haq} created exhaustive image descriptions in addition to the already existing short region-based descriptions. Doan et al.~\cite{2024-doan} utilized image-based generation for several purposes, such as description generation and relevant information extraction, to develop Vintern-1B, an efficient LMM for Vietnamese. Nath et al.~\cite{nath2022image} applied this approach for image caption generation in the low-resource Assamese language using an encoder-decoder framework that combines CNNs and RNNs to generate descriptions from images. Jiang et al.~\cite{jiang2024multimodal} expanded these approaches with multimodal seed data augmentation for the low-resource audio Latin Cuengh language, demonstrating how seed data can enhance intelligent recognition and comprehension of low-resource dialects.
Collectively, these studies demonstrate the versatility and effectiveness of synthetic data for tackling a diverse set of multimodal tasks in the context of LR languages.

An emerging approach that avoids both traditional back-translation or image-based methods consists in leveraging the outputs of Vision-Language Models (VLMs). Qu et al.~\cite{2024-qu} generated multilingual responses for image-text inputs, translated them into English, and compared them with trusted references to detect hallucinations. These automatically mined hallucination-aware pairs are then used for direct preference optimization \citep{2023-Rafailov-NeurIPS}, enabling scalable fine-tuning without manual annotations, especially useful for low-resource languages. 

Another innovative approach focuses on optimizing the composition of training data itself. Shukor et al.~\cite{shukor2025scaling_data} developed systematic methods to determine optimal domain weights for multimodal pre-training using scaling laws, validating their approach across Large Language Models, Native Multimodal Models, and Large Vision Models. This methodology provides principled alternatives to costly trial-and-error approaches for data mixture optimization, particularly valuable in resource-constrained settings, typical for low-resource language development.

\noindent\textbf{Data sovereignty concerns.} Synthetic data generation introduces risks beyond technical quality. Back-translation and LLM-based augmentation propagate source-language biases to target languages, potentially encoding cultural assumptions misaligned with target communities~\cite{navigli2023biases}. More critically, the CARE principles~\cite{carroll2020care} assert that Indigenous communities must retain authority over their linguistic data, a requirement that synthetic generation pipelines rarely accommodate. Evidence from S\'{a}mi language technology demonstrates the consequences: LLMs trained on available corpora without community oversight produce outputs that appear valid to non-speakers, but constitute nonsense to native speakers~\cite{wiechetek2024ethical}. We thus recommend that synthetic data pipelines incorporate community validation protocols and explicit data governance agreements prior to deployment.


\begin{figure*}[t!]
    \centering
    \includegraphics[width=1\textwidth]{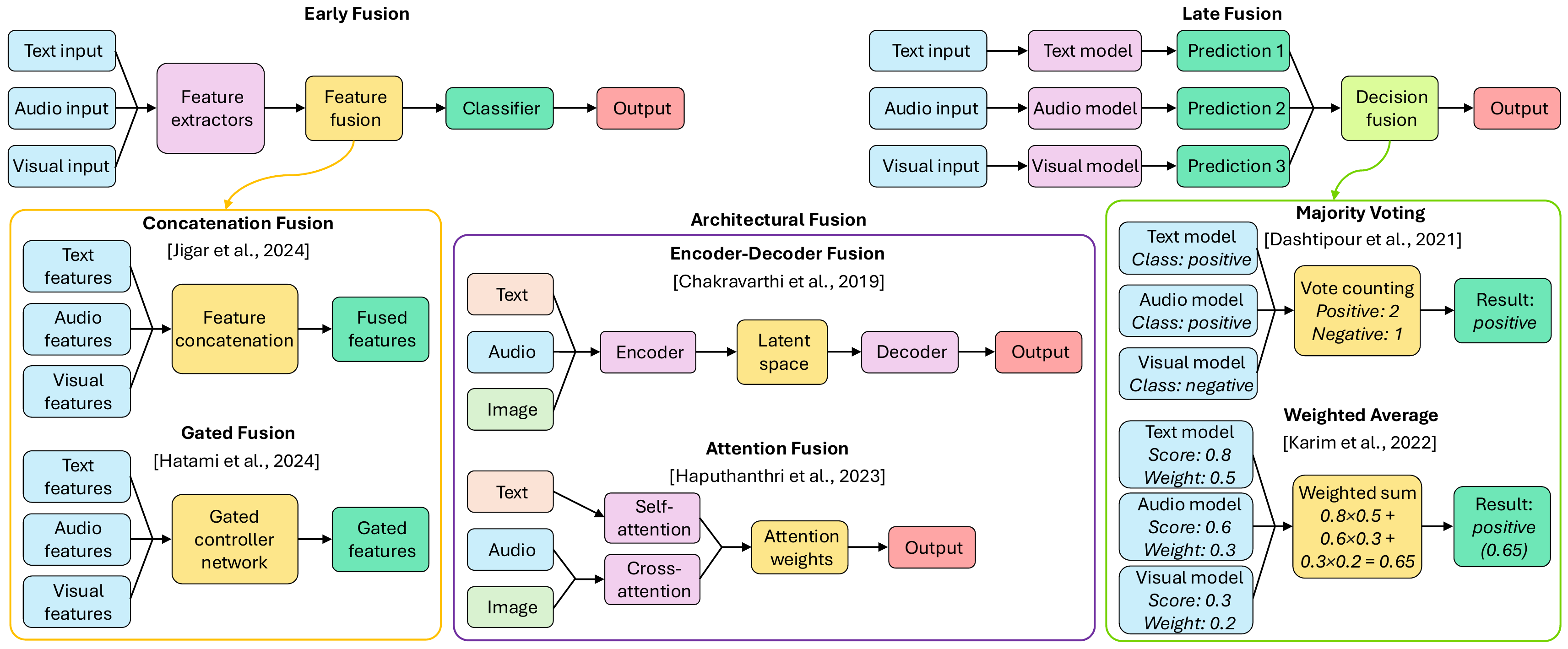}
    \vspace{-0.6cm}
    \caption{An overview of various fusion strategies employed in LMMs, categorized into early fusion, late fusion, and architectural fusion approaches. Early fusion combines features from different modalities (text, audio, and visual) using feature extractors and fusion techniques, before passing them to a classifier for the final output. Concatenation fusion directly concatenates features from different modalities, while gated fusion employs a gate controller network to regulate information flow between modalities. 
    Late fusion processes each modality using separate models, then combines their predictions using decision-level fusion methods, such as majority voting or weighted averaging. 
    Architectural fusion approaches, such as attention fusion and encoder-decoder fusion, provide more sophisticated methods for multimodal integration. Attention fusion leverages self-attention layers and learned attention weights to selectively focus on relevant features across modalities.}
    \label{fig:Fusion}
\end{figure*}

\section{Multimodal Fusion Techniques}
\label{sec_fusion}

Multimodal fusion refers to the process of combining information from different modalities (such as text, images, audio) to make more informed predictions or generate better outputs. Fusion can be seen as the ``meeting point'' where information from separate sensory channels comes together, similar to how humans integrate what they see, hear and smell to understand their environment. The choice of fusion strategy significantly impacts performance, especially in low-resource settings, where each modality might provide crucial complementary information that others lack. Below, we describe the primary approaches to fusion that represent different philosophies about when and how this integration should occur.

We identify three distinct types of fusion approaches employed in multimodal learning, categorized into early fusion, late fusion, and architectural fusion approaches. An overview of the different fusion strategies is provided in Figure \ref{fig:Fusion}. The diagram depicts the various ways in which textual, visual and auditory features can be combined at different stages to enable effective integration of multimodal information. In Table~\ref{tab:fusion_computational}, we provide a summary of computational requirements and efficiency trade-offs for a series of representative fusion approaches. We further discuss each fusion strategy independently, referring to the computational requirements and trade-offs along the way.

\noindent
\textbf{Early fusion.}
Early fusion, also known as feature-level fusion, involves combining features from different modalities at the input level before passing them through a unified model \citep{youcef2021arabic,2024-roken}. Early fusion can be conceptualized as ``combining ingredients before cooking'', i.e.~all modalities are mixed at the beginning of the processing pipeline. This allows the model to learn cross-modal interactions from the start, potentially capturing subtle relationships between modalities. 

In Persian sentiment analysis, Dashtipour et al.~\cite{2021-dashtipour} demonstrated the effectiveness of early fusion by combining acoustic, visual, and textual features through a context-aware framework, achieving 91.39\% accuracy with A+V+T concatenation. Similarly, Al-Azani et al.~\cite{2020-azani} showed that early fusion of textual, auditory, and visual modalities achieved over 94\% accuracy for Arabic sentiment analysis.

The shared task on Tamil and Malayalam multimodal sentiment analysis \citep{2023-premjith} also revealed that early fusion techniques are particularly effective for handling code-mixed content and cultural nuances specific to these languages \citep{kodali2025bytesizedllm}.

For Amharic hate speech detection in memes, Jigar et al.~\cite{2024-Jigar} employed concatenation, directly combining visual features from memes with textual features, achieving 75\% accuracy and demonstrating the effectiveness of this straightforward approach for LR languages \citep{2022-debele}. The integration of multimodal features through gating mechanisms has shown particular promise in LR scenarios, as demonstrated in English-to-Low-Resource translation tasks for Hindi, Malayalam, Bengali, and Hausa, where Hatami et al.~\cite{2024-Hatami} used gated fusion to selectively combine visual and textual information. This approach was further validated by Alalem et al.~\cite{2023-alalem} in their Audio-Text Fusion model for English and Egyptian Arabic, where they employed Group Gated Fusion to dynamically control the flow of information between modalities, achieving superior performance over traditional fusion methods. 

From a computational perspective, early fusion approaches such as Multi-Representative Fusion (MRF) \citep{2022-chauhan} demonstrate that competitive results can be achieved on consumer-grade hardware (GTX 1080Ti with 11\,GB VRAM), reaching 84.1\% accuracy on the ICT-MMMO dataset within 100 epochs. However, early fusion typically requires 2-3$\times$ more memory during training due to joint feature processing, and demands strict temporal alignment between modalities.

\begin{table*}[t]
\centering
\caption{Computational requirements and efficiency trade-offs for multimodal fusion techniques in low-resource settings. Bold values indicate configurations accessible for researchers with limited computational resources. A dash line indicates that the respective information is not specified in the original publication.}
\label{tab:fusion_computational}
\setlength\tabcolsep{0.3em}
\begin{adjustbox}{max width=\textwidth}
\begin{tabular}{@{}llccccp{5.8cm}@{}}
\toprule
\textbf{Method/Model} & \textbf{Fusion Type} & \textbf{GPU Req.} & \textbf{Training} & \textbf{Params} & \textbf{Performance} & \textbf{Key Trade-off} \\
\midrule
\multicolumn{7}{l}{\textit{Early Fusion Approaches}} \\
\midrule
MRF \citep{2022-chauhan} & Early & \textbf{1080Ti 11GB} & \textbf{100 ep.} & $\approx$50M & 84.1\% Acc & Noise-robust; needs multiple representations \\
ViT + mBERT \citep{faria2025sentimentformer} & Early & -- & \textbf{40 ep.} & $\approx$200M & 72.4\% Acc & High \#params; moderate accuracy \\
Swin + XLM-RoBERTa \citep{faria2025sentimentformer} & Early & -- & \textbf{40 ep.} & $\approx$280M & 75.8\% Acc & Best early fusion; heavier \\
A+V+T Concat \citep{2021-dashtipour} & Early & -- & -- & $\approx$30M & 91.4\% Acc & Simple; sync-sensitive \\
BiLSTM Multimodal \citep{2024-Jigar} & Early (Concat) & -- & \textbf{32-64 ep.} & $\approx$10M & 75.0\% Acc & Low \#params; limited complexity \\
\midrule
\multicolumn{7}{l}{\textit{Late Fusion Approaches}} \\
\midrule
XLM-R + DenseNet \citep{2022-karim} & Late & \textbf{GTX 1050} & \textbf{5-fold CV} & $\approx$400M & 83.0\% F1 & Best multimodal; high memory \\
MARBERTv2 + Ensemble \citep{2023-albalawi} & Late & -- & \textbf{100 ep.} & $\approx$180M & 85.6\% Acc & Robust to missing modalities \\
\midrule
\multicolumn{7}{l}{\textit{Intermediate / Architectural Fusion}} \\
\midrule
SentimentFormer \citep{faria2025sentimentformer} & Intermediate & -- & \textbf{30 ep.} & $\approx$220M & 79.0\% Acc & Best overall; balanced cost \\
AVTF-TBN \citep{zhang-2024} & Attention & RTX 3090 24GB & 300 ep. & $\approx$100M & 78.0\% F1 & High compute; medium accuracy \\
CNN-LSTM Tagalog \citep{deocampo2024lip} & Intermediate & -- & \textbf{12 h} & $\approx$20M & 89.5\% Acc & 25\% faster than A+V \\
\midrule
\multicolumn{7}{l}{\textit{Encoder--Decoder \& Advanced Fusion}} \\
\midrule
URSA (3D-CNN + BLSTM) \citep{2021-sehar} & Feature-level & -- & -- & 128+64 cells & 95.4\% Acc & Feature $>$ decision fusion \\
Feature-Extract \citep{2024-arifin} & Sep.+Merge & \textbf{T4 16GB} & -- & \textbf{8.48M} & 93.3\% Acc & Low \#params; specialized pipeline \\
\bottomrule
\end{tabular}
\end{adjustbox}
\end{table*}

\noindent
\textbf{Late fusion.} Late fusion, also known as decision-level fusion, combines predictions from separate modality-specific models at the decision stage rather than fusing features early in the pipeline \citep{deocampo2024lip,mamyrbayev2020multimodal}. Late fusion can be conceptualized as ``requesting multiple expert opinions and then voting on a final decision'' \citep{zhang-2024,2024-arifin}. In this approach, each modality is processed by its own specialized model, which becomes an expert in that particular type of data. Only after these individual experts have made their predictions are the results combined. This is particularly valuable when certain modalities might be missing or corrupted in real-world applications \citep{2023-elahi}. 


Two popular late fusion strategies are weighted averaging and majority voting. In weighted averaging, the predictions from different modalities are combined using a weighted sum, with weights determining the contribution of each modality to the final decision \citep{2024-rahman,2022-karim}. The weights can be uniform or learned to optimize performance. Majority voting employs gating mechanisms to control information flow between modalities and determine which modality should be emphasized \citep{das2022multi}. For example, Dashtipour et al.~\cite{2021-dashtipour} used gating networks to adaptively combine predictions from audio, visual and textual models based on their estimated reliability for Persian sentiment analysis. Their results showed that intelligent fusion using gates improved performance compared with simple averaging, highlighting the benefits of selective information integration in multimodal systems.

Late fusion strategies are especially suitable for resource-constrained environments due to their flexibility and lower memory requirements. Since each modality is processed by independent models, the system can continue functioning when one modality is unavailable, enabling graceful degradation with missing inputs \citep{2023-elahi,mamyrbayev2020multimodal}. For Arabic rumor detection, Albalawi et al.~\cite{2023-albalawi} achieved 83.83\% accuracy with late fusion (MARBERTv2 + VGG-19 ensemble), compared with 85.57\% for early fusion, demonstrating that the 1.7\% performance gap is often smaller than the computational cost savings. Late fusion also enables parallel training of modality-specific models, reducing wall-clock time by 25-40\% compared with end-to-end early fusion training \citep{deocampo2024lip}.

\noindent
\textbf{Architectural fusion.} 
Architectural fusion comprises more sophisticated integration methods that go beyond simple concatenation or averaging of features. Encoder-decoder fusion can be understood as a ``translation system'' between modalities, i.e.~information from each modality is first converted into a common ``language'' (shared representation space) by encoders, before being decoded into the final output \citep{faria2025sentimentformer,2024-roken}. This allows the model to find complex mappings between very different data types. For example, Chakravarthi et al.~\cite{2019-chakravarthi} employed an encoder-decoder framework with phonetic transcription to improve machine translation between Dravidian languages, while Sehar et al.~\cite{2021-sehar} utilized an encoder-decoder architecture to fuse audio, video and text features for Urdu sentiment analysis. Similarly, Meetei et al.~\cite{2023-meetei} showed that encoder-decoder fusion of correlated modalities can enhance translation quality for LR languages. The key advantage of encoder-decoder architectures is their ability to first encode input features from different modalities into a shared representation space before decoding them into the target output.

Attention-based fusion has also proven to be highly effective for multimodal integration \citep{ristea2023cascaded}. As shown by Haputhanthri et al.~\cite{haputhanthri-2023} for Sinhala sign language recognition, attention mechanisms allow the model to dynamically focus on the most relevant features across modalities. Yang et al.~\cite{2024-yang} successfully employed attention fusion for Mongolian sentiment analysis by combining features from audio, text and visual inputs. Zhang et al.~\cite{zhang-2024} demonstrated that attention-based fusion of multimodal data improves depression risk detection by allowing the model to attend to salient information across audio, video and text modalities. The ability of attention mechanisms to learn dynamic weights between modalities makes them particularly suitable for tasks requiring adaptive integration of complementary sources.

Intermediate and architectural fusion approaches offer a balance between performance and accessibility. SentimentFormer \citep{faria2025sentimentformer} achieves the highest Bangla meme accuracy (79.04\%) with only 30 epochs, outperforming both early (75.83\%) and late fusion (74.80\%) on the same dataset. At the high-resource end, attention-based models such as AVTF-TBN \citep{zhang-2024} require an RTX~3090 (24\,GB) and 300 epochs for clinical-grade depression detection accuracy.

\noindent
\textbf{Comparative analysis of fusion techniques.}
Each fusion approach presents distinct advantages and challenges in the context of LR languages \citep{youcef2021arabic,deocampo2024lip}. Early fusion enables deep interaction between modalities from the start, but can be computationally expensive and may suffer when one modality is noisy. Late fusion offers flexibility and robustness when modalities are missing, but may miss important cross-modal interactions \citep{mamyrbayev2020multimodal}. Architectural fusion approaches show promise in capturing complex relationships between modalities, but require careful tuning and substantial computational resources. A notable innovation in this space is the Multi-Representative Fusion (MRF) mechanism \citep{2022-chauhan}, which generates diverse representations for each modality and selectively chooses the best fusion via attention. This approach has shown particular promise in handling noisy inputs, achieving state-of-the-art performance on several LR sentiment analysis benchmarks.

\noindent\textbf{Handling noisy modalities.} A critical consideration for real-world deployment is robustness to corrupted modalities. The MRF mechanism \citep{2022-chauhan} addresses this by generating multiple diverse representations for each modality and using attention to select the most informative fusion. When acoustic features are corrupted, MRF automatically reduces their contribution (from approximately 15\% to $<$5\% of the final prediction), maintaining robust performance. However, MRF fails when all three modalities are simultaneously noisy for utterances critical to prediction. For Javanese emotion recognition \citep{2024-arifin}, separate modality processing achieves an accuracy of 93.32\% compared with 71.15\% for joint processing, specifically because independent processing minimizes interference when one channel contains noise.

\noindent\textbf{Architectural complexity considerations.} Our analysis suggests that the additional complexity of architectural fusion is justified in three cases: (1)~when cross-modal interactions are semantically rich and task-critical, as in Sinhala sign language recognition \citep{haputhanthri-2023} and Mongolian sentiment analysis \citep{2024-yang}, where temporal alignment between visual gestures and linguistic features requires learned attention weights; (2)~when modalities have different noise characteristics or information densities, as demonstrated for Urdu sentiment analysis where feature-level fusion (95.35\%) substantially outperformed decision-level fusion (91.23\%) \citep{2021-sehar}; and (3)~for clinical or safety-critical applications where prediction errors have serious consequences \citep{zhang-2024}. Conversely, for rapid prototyping or tasks where text modality dominates (e.g.~in Bengali hate speech detection, where a text-only XLM-RoBERTa achieves $F1=0.82$ vs.~$F1=0.83$ for the multimodal pipeline \citep{2022-karim}), simpler approaches may be preferable.

\begin{table}[t]
\centering
\caption{Performance comparison of early, late and intermediate fusion for low-resource languages. Best score on each row is highlighted in bold.} 
\label{tab:fusion_comparison}
\begin{adjustbox}{max width=\columnwidth}
\begin{tabular}{@{}lcccc@{}}
\toprule
\textbf{Language/Task} & \textbf{Early} & \textbf{Late} & \textbf{Intermediate} & \textbf{Best Strategy} \\
\midrule
Bangla Memes \cite{faria2025sentimentformer} & 75.83\% & 74.80\% & \textbf{79.04\%} & Intermediate \\
Arabic Rumors \cite{2023-albalawi} & \textbf{85.57\%} & 83.83\% & -- & Early \\
Urdu Sentiment \cite{2021-sehar} & \textbf{95.35\%} & 91.23\% & -- & Early \\
Javanese Emotion \cite{2024-arifin} & 71.15\% & -- & \textbf{93.32\%} & Separate Processing \\
Amharic Memes \cite{2024-Jigar} & \textbf{75.00\%} & -- & -- & Early \\
Persian Video \cite{2021-dashtipour} & \textbf{91.39\%} & 90.32\% & -- & Early \\
\bottomrule
\end{tabular}
\end{adjustbox}
\end{table}

\begin{table}[t]
\centering
\caption{Decision guide for selecting the fusion strategy based on constraints and requirements. \ding{51}\ding{51} = Strongly recommended, \ding{51} = Suitable, $\sim$ = Acceptable, \ding{55} = Not recommended. Based on empirical findings from \citep{2022-chauhan,faria2025sentimentformer,zhang-2024,mamyrbayev2020multimodal}.}
\label{tab:fusion_decision}
\small
\begin{tabular}{@{}p{5.5cm}ccc@{}}
\toprule
\textbf{Requirement/Constraint} & \textbf{Early} & \textbf{Late} & \textbf{Architectural} \\
\midrule
Missing modality robustness & \ding{55} & \ding{51}\ding{51} & \ding{51} \\
Noisy input handling & \ding{55} & \ding{51} & \ding{51}\ding{51} (MRF) \\
Low memory ($<$8GB VRAM) & \ding{55} & \ding{51}\ding{51} & $\sim$ \\
Fast training ($<$50 epochs) & $\sim$ & \ding{51}\ding{51} & $\sim$ \\
Maximum accuracy & \ding{51}\ding{51} & $\sim$ & \ding{51}\ding{51} \\
Cross-modal interactions & \ding{51}\ding{51} & \ding{55} & \ding{51}\ding{51} \\
Rapid prototyping & $\sim$ & \ding{51}\ding{51} & \ding{55} \\
Clinical/safety-critical & $\sim$ & \ding{55} & \ding{51}\ding{51} \\
\bottomrule
\end{tabular}
\end{table}

In Table~\ref{tab:fusion_comparison}, we present the performance of fusion strategies across different low-resource languages and tasks, while in Table~\ref{tab:fusion_decision}, we provide a decision guide based on specific constraints. Early fusion generally achieves the highest accuracy (e.g.~95.35\% for Urdu, 91.39\% for Persian video analytics), but the performance gap between strategies is often smaller than the computational cost difference. For researchers with limited computational resources (single GPU, $<$16GB VRAM), we recommend starting with lightweight early fusion models such as BiLSTM ($\approx$10M parameters) to establish baselines, before progressing to intermediate fusion with efficient architectures for improved performance.



\section{Visual Enhancement Techniques}
\label{sec_visual}

Visual enhancement techniques aim to improve MT quality by leveraging visual information to provide additional context and resolve ambiguities in the source text. These techniques broadly fall into two main categories: image-guided translation, which uses visual features to enhance the overall translation process, and visual disambiguation, which specifically focuses on resolving ambiguous words/phrases via visual context. 

\noindent\textbf{Image-guided translation.}
A promising direction for improving translation quality for LR languages is the use of image-guided translation approaches. Dutta Chowdhury et al.~\cite{2018-dutta} showed that augmenting neural MT systems with visual features extracted from a pre-trained CNN and integrated into an encoder-decoder architecture can improve translation quality, achieving a bilingual evaluation understudy (BLEU) score of 24.2 for Hindi to English translation. Building upon these ideas, Laskar et al.~\cite{2020-laskar, 2021-laskar-hindi} developed a multimodal neural MT system with a bidirectional RNN encoder and a doubly-attentive decoder for English-Hindi translation. Their system, which combines visual and textual features, and employs pre-trained word embeddings from monolingual data, outperforms a text-only baseline, achieving a BLEU score of 33.57 versus 27.75 on the test set.

Subsequent studies \citep{2023-gain,2022-shi,2023-meetei-cues,2023-meetei-hindi,2024-haq} have demonstrated the effective use of visual information for improving MT in LR settings, particularly for the English-Hindi language pair. Meetei et al.~\cite{2023-meetei-cues} proposed a video-guided multimodal MT framework that incorporates spatio-temporal video features, showing improvements of up to +4.2 BLEU over text-only baselines for English to Hindi translation, while Meetei et al.~\cite{2023-meetei-hindi} explored multimodal translation for news domain data, showing that ResNet-based image features outperform VGG-based features and improve BLEU scores by +1.8 points. Additionally, Shi et al.~\cite{2022-shi} explored different approaches for extracting and integrating image features using VGG and ResNet models, achieving a +3 BLEU improvement over text-only translation. Another contribution is presented by Gain et al.~\cite{2023-gain}, who showed how visual context enhances translation robustness under noisy conditions (e.g.~OCR errors), even when image relevance is reduced. Extending this line of work, Tayir et al.~\cite{2024-tayir} demonstrated that visual context can bridge structural gaps in distant language pairs, such as English-Uyghur, by introducing a visual masked language modeling approach for unsupervised multimodal MT. Similarly, Tayir et al.~\cite{2025-tayir} improved translation for the same language pairs by harnessing varying-granularity image features in low-resource settings.

More recently, Haq et al.~\cite{2024-haq} presented a context-aware transformer model that integrates visual features via BERT encoding, demonstrating consistent improvements over text-only baselines. In a related direction, Lekshmy et al.~\cite{2022-h-o} developed an English-Malayalam vision-aided translation system for visually impaired users, employing multimodal machine learning techniques to perform object recognition and generate translated descriptions in real-time.


Across all studies, qualitative analyses confirmed that visual cues are particularly beneficial for handling rare words and domain-specific terms, with both image and video modalities helping to resolve ambiguity and improve translation quality in LR scenarios.

\noindent
\textbf{Visual disambiguation.}
While image-guided translation aims to enhance overall translation quality by integrating visual context, the visual disambiguation techniques focus on task-specific ambiguities by grounding them in visual information. In this regard, studies revolving around the creation of Visual Genome datasets for LR languages, such as Hindi \citep{2019-parida}, Bengali \citep{2022-sen} and Hausa \citep{2022-abdulmumin}, have played a pivotal role in advancing visual disambiguation techniques. Building upon previous work, Parida et al.~\cite{2021-parida} explored this line of research by developing a multimodal NMT system for English-Bengali using object tags extracted from images as auxiliary input, while Nortje et al.~\cite{2024-nortje-2} introduced an innovative few-shot learning approach for visually-prompted keyword localization in Yoruba.

Several studies have investigated the use of visual features for disambiguation \citep{2021-jain,2022-jian,2024-kovath,2020-laskar,2022-laskar,2021-laskar-english-assamese}. For example, Jain et al.~\cite{2021-jain} highlighted the benefits of using visual features for disambiguation. Their model, called MURAL, shows strong performance on text-to-image retrieval tasks, where it manages to retrieve relevant images for ambiguous queries. This finding is also supported by the qualitative examples, where MURAL successfully disambiguates word senses based on visual context. In addition, Kovath et al.~\cite{2024-kovath} proposed a co-attention mechanism for Malayalam VQA that allows the model to jointly learn attention over both textual and visual inputs, demonstrating improved performance over baselines using only textual features.

\noindent
\textbf{Comparative analysis of visual enhancement techniques.}
Image-guided translation consistently demonstrates performance improvements over text-only baselines for LR languages, though effectiveness varies with dataset size and translation direction. These approaches excel at handling semantic ambiguities and culturally-specific concepts, but their success depends heavily on the quality of extracted visual features. A key limitation is the reliance on high-quality image-text pairs, which are often scarce for LR languages. While these techniques improve translation quality, they also introduce computational overheads. Future work should focus on developing more efficient visual feature extraction methods and better approaches for leveraging visual information with limited paired data.

\section{Cross-Modal Transfer Learning}
\label{sec_cross}

Cross-modal transfer learning represents a critical approach for LR languages, allowing models to harness knowledge from data-rich modalities or languages to improve performance in resource-constrained settings. Unlike traditional transfer learning, which operates within a single modality, cross-modal transfer must bridge the significant gap between different types of data representations. This is conceptually similar to how a person might use their understanding of written language to help learn a sign language, or how knowledge of one spoken language can facilitate learning another. In the context of low-resource languages, two primary transfer directions have emerged: modality transfer, which moves knowledge between different data types (e.g.~from text to speech), and language transfer, which leverages high-resource languages to improve performance in low-resource ones.

\noindent
\textbf{Modality transfer.}
Modality transfer addresses the challenge of transferring knowledge between different modalities to improve performance on low-resource tasks. This approach is particularly valuable when certain modalities have more abundant data than others. For example, text data is often easier to collect than paired speech data for many languages. The fundamental challenge lies in bridging the representational gap between modalities, since text operates in a discrete symbolic space, while speech and vision exist in continuous signal spaces with very different statistical properties. Successful modality transfer requires finding meaningful mappings between these different representational spaces.
A diversity of approaches has been used to achieve modality transfer. Chen et al.~\cite{2023-chen} proposed a progressive transfer learning strategy that leverages both general pre-training (Kinetics-400 for visual and CC25 for language) and domain-specific pre-training (sign-to-gloss translation) to bridge modalities for sign language translation. Amalas et al.~\cite{2024-amalas} introduced a data-driven approach to select source languages and demonstrated that multilingual pre-training outperforms monolingual pre-training for text-to-speech systems. Wu et al.~\cite{wu2022pairs} developed a captioning approach via multi-objective optimization that addresses the challenge of utilizing both triplet datasets (image, HR language, LR language) and large-scale paired datasets during training. Yeo et al.~\cite{2024-yeo} tackled LR visual speech recognition by using Whisper-based automatic transcriptions to generate training labels from unlabeled multilingual audio-visual data. For Arabic handwriting recognition, Bhatia et al.~\cite{bhatia2024qalammultimodalllm} employed modality transfer through an architecture combining SwinV2 for visual encoding and RoBERTa for text decoding, while Tran et al.~\cite{2024-tran-lavy} demonstrated successful modality transfer for Vietnamese through extensive pre-training of both vision and language components, combined with automated data curation methods. Notably, Onuoha et al.~\cite{2024-onuoha} challenged the assumptions about multimodal integration through their study of Igbo minimal pairs. Their findings show that native Igbo speakers can accurately distinguish minimal pairs through audio alone, suggesting that the benefits of cross-modal integration may be more relevant for non-native speakers.

\begin{table*}[t]
\centering
\caption{Overview of architectural innovations for low-resource multimodal learning. V = Vision, T = Text, A = Audio. Although Cycle-Attn is evaluated on EN and DE (high-resource), it is included as a key methodological reference. The authors simulated a low-resource scenario using the limited Multi30K dataset to demonstrate the efficacy of knowledge transfer from a rich monolingual corpus (EN) via cycle consistency constraints.}
\label{tab:arch_overview}
\setlength\tabcolsep{0.3em}
\begin{adjustbox}{max width=\textwidth}
\begin{tabular}{@{}llccclp{5.0cm}@{}}
\toprule
\textbf{Model/Method} & \textbf{Year} & \textbf{Languages} & \textbf{Modalities} & \textbf{Task} & \textbf{Approach Category} & \textbf{Key Innovation} \\
\midrule
Cycle-Attn \cite{2019-wu} & 2019 & EN, DE & V, T & Image Captioning & Translation+Alignment & Cycle consistency constraint for cross-lingual alignment \\
Multi-task Adversarial \cite{2022-mamta} & 2022 & EN, HI & T, A & Sentiment Analysis & Adversarial Learning & Cross-lingual transfer via shared embeddings \\
FEWVLM \cite{jin2022good} & 2022 & EN, HI & V, T & VL Understanding & Prompt Engineering & Few-shot prompting with moderate-size VLM \\
Amharic Captioning \cite{solomon2023amharic} & 2023 & Amharic & V, T & Image Captioning & Attention-based DNN & Visual attention + Bi-GRU decoder \\
Auxiliary CTC \cite{2023-chen} & 2023 & 102 langs & A, T & Multilingual ASR & CTC Conditioning & LID-conditioned auxiliary objectives \\
Sanskrit-Malayalam NMT \cite{rahul2023morphology} & 2022 & SA, ML & T, A & Machine Translation & Multimodal NMT & Morphology + WSD embedding fusion \\
XtremeCLIP \cite{2023-tang-xtremeclip} & 2023 & EN, HI & V, T & VL Understanding & Parameter-efficient & Prototype affinity matching (5-7K params) \\
LowCLIP \cite{2023-Asgarov} & 2024 & Azerbaijani & V, T & Image Retrieval & Efficiency-first & mBERT + lightweight image encoders \\
Yoruba ASR \cite{rahmon2024speech} & 2024 & EN, YO & T, A & Speech Recognition & Transfer Learning & MFCC-based acoustic modeling \\
Llama 3 \cite{lamma} & 2024 & 200 langs & V, T & General Multimodal & Foundation Model & Native multimodal MoE architecture \\
DeepSeek-V3 \cite{DeepSeek} & 2024 & 14 langs & V, T & General Multimodal & MoE Architecture & MLA + FP8 training efficiency \\
Claude 4 \cite{anthropic2025claude} & 2025 & 15 langs & V, T & General Multimodal & Foundation Model & Dual-mode reasoning operation \\
Apple AFM \cite{DBLP:journals/corr/abs-2407-21075} & 2024 & 16 langs & V, T & On-device/Server & Distillation+QAT & 2-bit quantization for edge deployment \\
MMaDA \cite{MMaDA} & 2025 & 60+ langs & V, T & Multimodal Diffusion & Unified Diffusion & Discrete diffusion language modeling \\
MixLoRA \cite{Mixture_of_LoRA} & 2024 & 15 langs & V, T & Instruction Tuning & Dynamic PEFT & Conditional mixture routing for adaptation \\
\bottomrule
\end{tabular}
\end{adjustbox}
\end{table*}

\noindent
\textbf{Language transfer.}
Language transfer is an approach to harness knowledge from HR languages to improve model performance on LR languages. Recent work demonstrated several effective strategies. For instance, Wang et al.~\cite{2023-wang-adapting} adapted MDETR to new languages by using adapters and code-switching without relying on MT data. Cheema et al.~\cite{cheema2024adapting} presented ViLanOCR, a novel approach that adapts multilingual vision-language transformers for low-resource Urdu optical character recognition by leveraging the Swin encoder and mBART-50 decoder. Kim et al.~\cite{2023-kim} focused on learning general speech knowledge from English for lip reading, and combining it with language-specific audio features. Aruna Gladys et al.~\cite{vetriselvi2024sentiment} proposed a multimodal representation learning framework that uses cross-lingual transfer learning to analyze sentiment in LR language datasets, demonstrating significant performance improvements for Tamil language sentiment analysis. Chen et al.~\cite{2023-wchen} improved multilingual ASR by conditioning models on language identity predictions from early layers to enhance performance across numerous languages. dos Santos et al.~\cite{2023-santos} proposed to use data augmentation and contrastive learning to improve multilingual contrastive language-image pre-training (CLIP) models for LR languages. Nortje et al.~\cite{2024-nortje} showed that initializing a Yoruba few-shot word learning model with weights from an English speech-image model substantially improves performance. These approaches share the common theme of transferring learned representations and knowledge from HR languages (typically English), while developing techniques to adapt and fine-tune models for target LR languages.

\begin{table*}[t]
\centering
\caption{Computational requirements for architectural innovations. A dash line indicates that the respective information was not reported in the original paper.}\label{tab:computational_efficiency}
\begin{adjustbox}{max width=\textwidth}
\begin{tabular}{@{}lccccc@{}}
\toprule
\textbf{Model} &
\makecell{\textbf{Trainable}\\\textbf{Params}} &
\makecell{\textbf{Total}\\\textbf{Params}} &
\makecell{\textbf{Training}\\\textbf{Duration}} &
\makecell{\textbf{Hardware}\\\textbf{(per paper)}} &
\makecell{\textbf{Training}\\\textbf{Data}} \\
\midrule
\multicolumn{6}{l}{\textit{Parameter-Efficient Vision-Language Methods}} \\
\midrule

XtremeCLIP \cite{2023-tang-xtremeclip} & \textbf{5-7K} & 149M & 20-60 min & 1$\times$ A100 & 2K-10K samples \\
LowCLIP \cite{2023-Asgarov} & 192M & 192M & 37 hours & 1$\times$ T4 & 500K+ captions \\
FEWVLM$_{\text{base}}$ \cite{jin2022good} & 224M & 224M & 30 epochs & -- & Few-shot (16 ex.) \\
FEWVLM$_{\text{large}}$ \cite{jin2022good} & 740M & 740M & 30 epochs & -- & Few-shot (16 ex.) \\

\midrule
\multicolumn{6}{l}{\textit{Language-Specific Architectures}} \\
\midrule

Amharic Caption \cite{solomon2023amharic} & -- & -- & 35 epochs & -- & 8K images \\
Cycle-Attn \cite{2019-wu} & -- & -- & 50 epochs & -- & 30K pairs \\
\midrule
\multicolumn{6}{l}{\textit{Foundation Models (for reference)}} \\
\midrule

Llama 3 405B \cite{lamma} & 405B & 405B & $3.8\times10^{25}$ FLOPs & 16K$\times$ H100 & 15.6T tokens \\
DeepSeek-V3 \cite{DeepSeek} & 37B active & 671B & 2.788M H800 hours & 2048$\times$ H800 & 14.8T tokens \\
\bottomrule
\end{tabular}
\end{adjustbox}
\vspace{1mm}
\end{table*}

The effectiveness of language transfer methods varies significantly based on linguistic similarity, writing systems, and cultural context. For instance, transfer between closely related languages (such as Spanish to Portuguese) typically outperforms transfer between distant language families (such as English to Tamil). The methods described above demonstrated different approaches to this challenge: Wang et al.~\cite{2023-wang-adapting} focused on architecture adaptation through adapters, while Kim et al.~\cite{2023-kim} emphasized feature-level knowledge transfer. Meanwhile, Nortje et al.~\cite{2024-nortje} showed that even initialization from a different language can provide substantial benefits. For practitioners working with specific low-resource languages, the choice between these approaches should consider both linguistic factors and computational constraints.

\section{Architectural Innovations}
\label{sec_arch}
\begin{table*}[t!]
\centering
\caption{Performance metrics for low-resource multimodal architectures. All values are extracted directly from source papers. Baseline methods and improvement calculations are specified for reproducibility. Full FT = full fine-tuning; Aug. = augmentation; -- = not applicable or not reported.}
\label{tab:performance_metrics}
\begin{adjustbox}{max width=\textwidth}
\begin{tabular}{@{}llllccc@{}}
\toprule
\textbf{Model} & \textbf{Task} & \textbf{Dataset} & \textbf{Metric} & \textbf{Score} & \textbf{Baseline} & \textbf{$\Delta$} \\
\midrule
\multicolumn{7}{l}{\textit{Parameter-Efficient Vision-Language Methods}} \\
\midrule
\multirow{3}{*}{XtremeCLIP \cite{2023-tang-xtremeclip}} 
  & Visual Entailment & SNLI-VE (10K samples) & Accuracy & 62.06\% & 51.10\% (Full FT) & +21.4\% \\
  & Visual QA & VQA v2 (10K samples) & Accuracy & 59.21\% & 54.10\% (Full FT) & +9.4\% \\
  & Image Classification & FGVC (16-shot) & Accuracy & 48.30\% & 28.14\% (Full FT) & +71.6\% \\
\midrule
\multirow{2}{*}{LowCLIP \cite{2023-Asgarov}} 
  & \multirow{2}{*}{Image Retrieval} & MSCOCO (AZ) & mAP & 0.80 & 0.70 (Base Loss) & +14.3\% \\
  & & Flickr30k (AZ) & mAP & 0.87 & 0.84 (No Aug.) & +3.6\% \\
\midrule
\multirow{2}{*}{FEWVLM$_{\text{large}}$ \cite{jin2022good}} 
  & \multirow{2}{*}{Visual QA} & VQAv2 (few-shot) & Accuracy & 51.1\% & 38.2\% (Frozen 7B) & +33.8\% \\
  & & OK-VQA (few-shot) & Accuracy & 23.1\% & 12.6\% (Frozen 7B) & +83.3\% \\
\midrule
\multicolumn{7}{l}{\textit{Language-Specific Architectures}} \\
\midrule
\multirow{2}{*}{Amharic Captioning \cite{solomon2023amharic}} 
  & \multirow{2}{*}{Image Captioning} & Flickr8k (AM) & 4-gram BLEU & 38.8 & 28.5 (CNN-GRU) & +36.1\% \\
  & & BNATURE (AM) & 4-gram BLEU & 42.7 & 16.4 (CNN-GRU) & +160.4\% \\
\midrule
\multirow{2}{*}{Cycle-Attn \cite{2019-wu}} 
  & \multirow{2}{*}{Image Captioning} & \multirow{2}{*}{Multi30K (DE)} & CIDEr & 43.78 & 42.91 (Dual-Attn+) & +2.0\% \\
  & & & BLEU-4 & 5.71 & 5.54 (Dual-Attn+) & +3.1\% \\
\midrule
\multicolumn{7}{l}{\textit{Foundation Models (for reference)}} \\
\midrule
Llama 3 405B \cite{lamma} & General & MMLU (5-shot) & Accuracy & 87.3\% & -- & -- \\
DeepSeek-V3 \cite{DeepSeek} & General & MMLU-Pro (5-shot CoT) & Accuracy & 75.9\% & -- & -- \\
\bottomrule
\end{tabular}
\end{adjustbox}
\end{table*}

Architectural innovations for low-resource multimodal learning focus on designing model structures that can effectively leverage limited data while maintaining reasonable computational requirements. The fundamental challenge lies in balancing model capacity (ability to learn complex patterns) with sample efficiency (ability to learn from limited examples). While simply scaling down large models designed for high-resource settings is one approach, the most successful innovations in this space incorporate architectural elements specifically designed to address the constraints of low-resource scenarios. These innovations generally fall into three categories: (1) efficiency-focused adaptations of existing architectures, (2) parameter-efficient fine-tuning methods, and (3) novel architectures designed specifically for low-resource multimodal learning. In Table~\ref{tab:arch_overview}, we provide a systematic overview of these architectural innovations, categorized by approach type, supported modalities, and target tasks. Tables~\ref{tab:computational_efficiency} and~\ref{tab:performance_metrics} complement this overview with quantitative analyses of computational requirements and empirical performance, enabling direct comparison across methods with varying resource constraints.

Some recent architectural innovations in the context of LR languages have focused on adapting the CLIP architecture \citep{radford2021learning}. One such example is the LowCLIP model \citep{2023-Asgarov}, which replaces the original text encoder trained primarily on English text with a multilingual BERT (mBERT). The authors evaluated various lightweight image encoders, such as EfficientNet-B0 and Tiny Swin Transformer, for a more computationally efficient approach, while also targeting LR languages like Azerbaijani. To compensate for the lighter architecture and the scarcity of image-text pairs in Azerbaijani, LowCLIP leveraged synthetic data generation via MT for text features, and image augmentation techniques, such as crop and rotation, for image features. In contrast, XtremeCLIP \citep{2023-tang-xtremeclip} took a different approach, in which the authors introduced a parameter-efficient method that only tunes a small prototype matrix, while keeping the visual and text encoders frozen. Their model also employs contrastive learning to provide additional supervisory signals in LR settings. Collectively, these efforts extend the applicability of CLIP to multimodal image retrieval tasks.

\begin{table}[t]
\centering
\caption{Image encoder performance comparison for low-resource image retrieval. Results are taken from LowCLIP \cite{2023-Asgarov}.}
\label{tab:image_encoders}
\begin{adjustbox}{max width=\textwidth}
\begin{tabular}{@{}lcccccc@{}}
\toprule
\textbf{Image Encoder} & \textbf{Params} & \textbf{GFLOPs} & \textbf{Size (MB)} & \multicolumn{3}{c}{\textbf{mAP}} \\
\cmidrule(lr){5-7}
& & & & COCO & Flickr8k & Flickr30k \\
\midrule
ResNet-50 & 25.6M & 4.09 & 97.8 & 0.80 & 0.76 & 0.73 \\
EfficientNet-B0 & 5.29M & 0.39 & 20.5 & \textbf{0.81} & 0.85 & \textbf{0.87} \\
ViT-Base & 86.6M & 17.56 & 330.3 & 0.71 & 0.80 & 0.70 \\
Swin-Tiny & 28.3M & 4.49 & 108.2 & 0.80 & \textbf{0.84} & 0.79 \\
\bottomrule
\end{tabular}
\end{adjustbox}
\end{table}

Approaches to adapting CLIP for LR settings illustrate different design philosophies. LowCLIP takes an efficiency-first approach, focusing on reducing both the model size and data requirements through lighter architectures and extensive data augmentation. In contrast, XtremeCLIP maintains most of the original model capacity, but introduces parameter-efficient tuning to learn a small set of adaptable weights. This trade-off between model capacity and training efficiency represents a key consideration for researchers working in low-resource settings, where both data and computational resources may be constrained. The choice between these approaches depends on the specific constraints of the application scenario, e.g.~LowCLIP may be more suitable for deployment on edge devices or in settings with extremely limited data, while XtremeCLIP might be preferred when maintaining representation power for complex tasks is crucial. 
As shown in Table~\ref{tab:image_encoders}, EfficientNet-B0 achieves competitive retrieval performance (an mAP of 0.87 on Flickr30k), while requiring 16$\times$ fewer parameters and 45$\times$ fewer FLOPs than ViT-Base. The choice between these approaches depends on deployment constraints: LowCLIP suits scenarios requiring end-to-end retraining with domain-specific data, while XtremeCLIP is preferable when rapid adaptation with minimal computational overhead is essential.

Another approach for multimodality in the context of LR languages is introduced by Wu et al.~\cite{2019-wu}. The approach combines two existing methods, a translation-based one and an alignment-based one, into a unified architecture to improve image captioning. The framework employs a model that first generates high-quality English captions, which are then used together with the images to produce captions in the LR language. The model achieves a fine-grained alignment between visual elements and captions in both languages via a cycle-consistency constraint, outperforming existing methods on standard metrics.

\begin{table}[t!]
\centering
\caption{Comparison of parameter-efficient fine-tuning methods for low-resource vision-language understanding. VE = Visual Entailment, VQA = Visual Question Answering, IC = Image Classification. Results are taken from XtremeCLIP \cite{2023-tang-xtremeclip}.}
\label{tab:peft_comparison}
\begin{adjustbox}{max width=\columnwidth}
\begin{tabular}{@{}lcccccc@{}}
\toprule
\textbf{Method} & \textbf{Trainable Params} & \textbf{VE} & \textbf{VQA} & \textbf{IC} & \textbf{Avg.} & \textbf{Training Time} \\
\midrule
Zero-shot & 0 & 33.74 & 52.03 & 39.17 & 42.89 & -- \\
Full fine-tuning & 149M & 51.10 & 54.10 & 28.14 & 51.12 & hours \\
LLRD & 149M & 57.23 & 53.88 & 31.36 & 53.60 & hours \\
BitFit & 176-178K & 59.56 & 54.72 & 41.61 & 55.66 & minutes \\
BiNor & 208-210K & 59.54 & 54.75 & 41.73 & 55.67 & minutes \\
CLIP-Adapter & 131-262K & 59.21 & 54.21 & 44.88 & 55.45 & minutes \\
Tip-Adapter & 5-10M & 59.67 & 54.70 & 45.12 & 55.62 & minutes \\
\textbf{XtremeCLIP} & \textbf{5-7K} & \textbf{62.06} & \textbf{59.21} & \textbf{48.30} & \textbf{57.73} & \textbf{20 minutes} \\
\midrule
LoRA ($r=4$) & $\approx$4K & -- & -- & -- & 65.39 & minutes \\
LoRA ($r=16$) & $\approx$16K & -- & -- & -- & 65.50 & minutes \\
\textbf{MixLoRA ($E=16$)} & $\approx$8K/layer & -- & -- & -- & \textbf{67.17} & hours \\
\bottomrule
\end{tabular}
\end{adjustbox}
\end{table}

Jin et al.~\cite{jin2022good} introduced FEWVLM, showing that careful prompt engineering and efficient architectural design can achieve strong performance in the context of LMM usage with either little data or computational needs. They managed to develop a moderate-size VLM that combines a sequence-to-sequence transformer with prefix language modeling and masked language modeling, introducing effective prompt engineering approaches for visual-language tasks in the LR setting. Notably, FEWVLM outperforms Frozen \citep{10.5555/3540261.3540277}, a model which is 31$\times$ larger. In turn, Frozen achieves comparable results with PICa \citep{yang2022empirical}, which is 246$\times$ larger. These results demonstrate that an effective design can compensate for model size. Building on parameter-efficient approaches, Shen et al.~\cite{Mixture_of_LoRA} introduced Conditional Mixture of LoRA (MixLoRA) for multimodal instruction tuning, which dynamically constructs adaptation matrices tailored to each input instance, addressing task interference challenges in multimodal scenarios. For specific language pairs, Laskar et al.~\cite{2023-laskar} proposed a transliteration-based phrase augmentation approach for English-Assamese translation, which allows their model to share sub-word level information, and provides better word alignment through phrase pairs. In Table~\ref{tab:peft_comparison}, we quantify the size vs.~performance trade-off across these methods. XtremeCLIP achieves the highest average accuracy (57.73\%) across visual entailment, VQA, and image classification benchmarks, while training only 5-7K parameters, compared with 149M for full fine-tuning. This demonstrates that task reformulation as prototype affinity matching can outperform conventional fine-tuning, while using less than 21,000$\times$ fewer trainable parameters. FEWVLM$_{\text{large}}$ (740M parameters) achieves 51.1\% on VQAv2, surpassing the 7B-parameter Frozen model (38.2\%) by 33.8\%, validating the hypothesis that architectural efficiency can compensate for raw model scale. MixLoRA further improves upon standard LoRA by 8.3\% on the MME benchmark through its conditional mixture routing mechanism, which dynamically selects expert combinations based on input characteristics.

Foundation models represent a qualitatively different design point, prioritizing broad capability over resource efficiency. We include them here to contextualize the computational differences that shape research accessibility. Dubey et al.~\cite{lamma} introduced the Llama~3 series with models ranging from 8B to 405B parameters, officially supporting 8 languages (English, German, French, Italian, Portuguese, Hindi, Spanish, and Thai), with experimental multilingual capabilities on a broader set via the speech interface (34 languages). Llama~3 multimodal extensions for image, video, and speech understanding were described in their technical report, but remain under development and have not been publicly released along with the paper. In a similar endeavor, Liu et al.~\cite{DeepSeek} presented DeepSeek-V3, a 671B-parameter MoE language model (37B active parameters per token) with multi-head latent attention (MLA) and FP8 mixed-precision training. It is important to note that DeepSeek-V3 is a text-only language model without native vision or audio capabilities. However, we include it to put efficient training strategies into perspective (DeepSeek-V3 requires 2.788M H800 GPU-hours, costing approximately \$5.6M) and better inform future multimodal model development. For edge deployment scenarios, Gunter et al.~\cite{DBLP:journals/corr/abs-2407-21075} introduced Apple Intelligence Foundation Models with a novel Parallel-Track MoE architecture optimized for on-device processing, supporting 16 languages with 2-bit quantization-aware training. The prevalence of MoE architectures in these recent developments demonstrates the effectiveness of expert-based scaling for multimodal tasks, as also observed by Mu et al.~\cite{DBLP:journals/corr/abs-2503-07137}. Additionally, Yang et al.~\cite{MMaDA} proposed a unified diffusion architecture that combines multimodal understanding with generation capabilities, offering new perspectives on architectural design for LR contexts.

\begin{table*}[t!]
\centering
\caption{Design strategies and trade-offs in multimodal architectures for low-resource settings. Core strategies are grouped by methodological approach.}
\label{tab:design_tradeoffs}
\begin{adjustbox}{max width=\textwidth}
\begin{tabular}{@{}lp{6cm}p{6cm}p{6cm}@{}}
\toprule
\textbf{Model} & \textbf{Core Strategy} & \textbf{Advantages} & \textbf{Constraints} \\
\midrule
\multicolumn{4}{l}{\textit{Parameter-Efficient Adaptation}} \\
\midrule
XtremeCLIP \cite{2023-tang-xtremeclip} & 
  Prototype affinity matching with frozen CLIP encoders; contrastive learning for supervision & 
  21,000$\times$ less parameters vs.~full fine-tuning; 20 min training on one A100; edge-deployable & 
  Task performance bounded by frozen backbone capacity; requires labeled prototype examples \\
\addlinespace
LowCLIP \cite{2023-Asgarov} & 
  Lightweight image encoders (EfficientNet-B0) with mBERT; synthetic data via MT & 
  Trainable on consumer GPU (T4); open-source; 37 hours total training & 
  Performance depends on MT quality; cross-domain generalization gap observed \\
\addlinespace
FEWVLM \cite{jin2022good} & 
  Seq2seq with PrefixLM + MaskedLM; prompt-based few-shot learning & 
  Outperforms 31$\times$ larger Frozen model; comparable with 246$\times$ larger PICa & 
  Zero-shot performance sensitive to prompt wording; task-specific prompt engineering required \\
\addlinespace
MixLoRA \cite{Mixture_of_LoRA} & 
  Conditional mixture of LoRA experts; input-dependent routing & 
  Reduces task interference in multi-task settings; 8.3\% gain over standard LoRA on MME & 
  Routing computation overhead; requires careful expert initialization \\
\midrule
\multicolumn{4}{l}{\textit{Cross-Lingual Transfer}} \\
\midrule
Cycle-Attn \cite{2019-wu} & 
  Translation + alignment hybrid with cycle consistency constraint & 
  Fine-grained visual-textual alignment; leverages English captioning supervision & 
  Requires pre-trained English captioner; limited to language pairs with English pivot \\
\addlinespace
Amharic Captioning \cite{solomon2023amharic} & 
  Inception-v3 encoder + Bi-GRU decoder with visual attention & 
  End-to-end trainable; interpretable attention weights; significant BLEU increase on BNATURE & 
  Requires translated Flickr8k data; architecture not tested on other LR languages \\
\midrule
\multicolumn{4}{l}{\textit{Multilingual Speech}} \\
\midrule
Auxiliary CTC \cite{2023-chen} & 
  LID-conditioned auxiliary objectives on Whisper encoder & 
  Scales to 102 languages; 28\% relative CER reduction on FLEURS & 
  Requires pre-extracted Whisper features; multi-stage training pipeline \\
\midrule
\multicolumn{4}{l}{\textit{Foundation Models (for reference)}} \\
\midrule
Llama 3 \cite{lamma} & 
  Dense Transformer (405B params); multimodal extensions under development & 
  Strong zero-shot; 8 officially supported languages; open weights & 
  $3.8 \times 10^{25}$ FLOPs pre-training; multimodal capabilities not yet released \\
\addlinespace
DeepSeek-V3 \cite{DeepSeek} & 
  MoE with MLA (671B total, 37B active); FP8 mixed-precision training & 
  2.788M H800 GPU hours (\$5.6M); competitive with GPT-4o on benchmarks & 
  Text-only model; no native vision/audio; requires 2048$\times$H800 cluster \\
\addlinespace
Apple AFM \cite{DBLP:journals/corr/abs-2407-21075} & 
  On-device ($\approx$3B) with 2-bit QAT; server PT-MoE architecture & 
  Edge-deployable; 16 languages; image understanding capability & 
  Proprietary; Apple ecosystem only; version-specific adapters \\
\bottomrule
\end{tabular}
\end{adjustbox}
\end{table*}

For specific language families and modality combinations, several innovative architectures have been proposed. Solomon et al.~\cite{solomon2023amharic} developed a hybridized attention-based deep neural network for Amharic language image captioning, combining a CNN encoder with visual attention mechanisms and a bidirectional GRU decoder, achieving significant improvements in terms of BLEU. Rahul et al.~\cite{rahul2023morphology} introduced a multimodal neural machine translation system between Sanskrit and Malayalam, which embeds morphology and word sense disambiguation awareness. It utilizes both textual and speech modalities via a two-level fusion approach of transform-based feature vectors. For African languages, Rahmon et al.~\cite{rahmon2024speech} presented a speech recognition model for Yoruba that employs acoustic and language modeling with sequential MFCC features, achieving 83\% accuracy in speech-to-text conversion. For sentiment analysis, Mamta et al.~\cite{2022-mamta} explored multilingual, multi-task and adversarial learning approaches to transfer knowledge from HR languages to LR scenarios, leveraging shared semantic spaces through cross-lingual word embeddings.
For Arabic, Alwajih et al.~\cite{2024-Alwajih} introduced Peacock, a comprehensive family of LMMs with strong vision and language capabilities, alongside Henna, a benchmark for evaluating culturally-aware Arabic LMMs, further helping to bridge the gap between high-resource and low-resource languages, while addressing unique linguistic and cultural characteristics.

In Table~\ref{tab:design_tradeoffs}, we synthesize the design trade-offs across architectural approaches, organized by methodological strategy. Three principal patterns emerge from our analysis. First, \textit{parameter-efficient adaptation} methods (XtremeCLIP, LowCLIP, FEWVLM, MixLoRA) achieve competitive performance, while reducing trainable parameters by 3-5 orders of magnitude compared with full fine-tuning, making the corresponding models accessible to researchers with limited computational resources. Second, \textit{cross-lingual transfer} approaches (Cycle-Attn, Amharic Captioning) effectively leverage high-resource language supervision, typically English, to bootstrap performance in target languages. However, this creates structural dependency on pivot language quality and availability. Third, foundation models occupy a distinct design regime. Since Llama 3 requires $3.8 \times 10^{25}$ FLOPs and DeepSeek-V3 consumes 2.788M H800 GPU-hours for pre-training, these models remain inaccessible to most research groups focused on low-resource languages. The practical implication is that parameter-efficient methods currently offer the most viable path for researchers operating under resource constraints, while foundation models may serve as upstream components for transfer learning when API access or pre-trained weights are available.

The computational requirements documented in Table~\ref{tab:computational_efficiency} reveal a structural divide with sociolinguistic implications. While parameter-efficient methods like XtremeCLIP (5-7K parameters, 20 minutes on one GPU) remain accessible, foundation models require resource-intensive infrastructure (Llama~3 consumes $3.8 \times 10^{25}$ FLOPs across 16K H100 GPUs; DeepSeek-V3 requires 2.79M H800 GPU hours with an estimated cost of \$5.6M). This asymmetry matters because LLMs exhibit systematic bias in knowledge acquisition. Indeed, new knowledge is learned less efficiently in LR languages, transfers less effectively to them, and is overwritten more easily by HR language information~\cite{wu2025inequalities}. The implication is that scaling alone is not sufficient to achieve equity. Therefore, architectural innovations must explicitly counteract these biases.

Federated learning offers a technical framework aligned with data sovereignty principles, enabling collaborative training without data centralization~\cite{lin2022fednlp}. Recent work demonstrates feasibility for multilingual LR settings. For example, federated prompt tuning achieves competitive performance while preserving data locality~\cite{chen2025breaking}, and differential privacy integration protects against gradient inversion attacks~\cite{yang2025dpfpl}. For multimodal LR applications, federated approaches could enable geographically-distributed language communities to collaboratively improve models without ceding control over culturally-sensitive audiovisual data.

\section{Evaluation Challenges}
\label{sec_eval}

Evaluation remains one of the most underdeveloped aspects of research on LMMs for LR languages. While the field has made significant strides in dataset creation, fusion strategies, and architectural innovations, the ways for measuring success have not kept pace. The lack of consistent and culturally-grounded evaluation protocols severely hampers the ability of researchers to compare models, reproduce findings, or interpret results in real-world contexts.

\noindent\textbf{Limitations of standard metrics across cultural contexts.} Most evaluation pipelines for LR multimodal models rely on automatic metrics originally designed for high-resource and predominantly Western-centric settings. Metrics such as BLEU, ROUGE, accuracy, and F1 implicitly assume that reference annotations reflect shared cultural, visual, and linguistic grounding. This assumption frequently fails in low-resource contexts.

One such case can be observed in multimodal tasks such as visual question answering, image captioning, and meme understanding, where the visual content itself often encodes culturally-specific assumptions regarding object salience, social roles, or everyday activities. For instance, a model trained primarily on Western image datasets may fail to recognize culturally significant objects (e.g.~traditional clothing, local foods, religious symbols, etc.) that are common in LR language contexts. When benchmarks are translated or minimally adapted from high-resource languages, models may achieve high lexical overlap with reference answers while still producing outputs that are culturally inappropriate, semantically misleading, or pragmatically implausible for native speakers. These issues are further exacerbated in knowledge-intensive evaluations derived from English-centric benchmarks. For example, questions about local festivals, historical events, or social customs require cultural context that translation alone cannot provide. As a result, standard metrics may overestimate progress or mask systematic failures that are only visible through culturally-grounded evaluation.

\noindent\textbf{Dataset heterogeneity and comparability issues.} A second major challenge concerns dataset heterogeneity, as existing studies evaluate multimodal models on datasets with widely distinct characteristics and assumptions. Many studies rely on translated versions of high-resource benchmarks, such as extensions of Multi30K for Ukrainian \cite{2023-saichyshyna} or Visual Genome variants for Bengali \cite{2022-sen}, Hausa \cite{2022-abdulmumin}, and Hindi \cite{2019-parida}. While translation-based approaches enable rapid benchmark construction, they often introduce Western cultural biases and may fail to reflect authentic language use or visual grounding in target communities. In contrast, newly introduced language-specific datasets, such as DravidianMultiModality \cite{2021-Chakravarthi}, RoMemes \cite{Pais2024RoMemesAM}, and ArabSign \cite{2022-hamzah}, better capture genuine linguistic and cultural phenomena, but typically suffer from limited coverage, non-standardized annotation protocols, and heterogeneous quality control practices, making cross-study comparison difficult. As a result, performance improvements reported across such heterogeneous evaluation settings are often not directly comparable.

\noindent\textbf{Recommendations for fair evaluation practices.} Based on our analysis, we propose the following recommendations for evaluation in LR multimodal research:
\begin{itemize}
    \item \textit{Report multiple metrics.} Studies should report multiple complementary metrics that capture different aspects of performance. In the context of machine translation tasks, researchers should report BLEU alongside COMET, or human evaluation scores. Another example in the context of VQA tasks, exact-match accuracy should be accompanied by relaxed matching that accounts for morphological variants and, when possible, human judgment of answer correctness.
    
    \item \textit{Perform culturally-grounded human evaluation.} In addition to a diverse set of automated metrics, we believe that human evaluation conducted by native speakers from the target language community also plays a crucial role. Evaluators should assess whether outputs sound natural to native speakers, whether they are culturally appropriate, whether they convey the intended meaning accurately, and (for VQA) whether answers are semantically correct, even if worded differently from the reference.

    \item \textit{Develop and use standardized benchmarks.} The field needs publicly available test sets for LR multimodal evaluation, following examples like SEACrowd \cite{2024-lovenia} for Southeast Asian languages and CreoleVal \cite{lent-etal-2024-creoleval} for Creole languages. Such benchmarks should cover different task types (VQA, captioning, translation, classification), include culturally-accurate content created together with language communities, provide multiple correct answers to account for natural variation, and document how data was labeled.

    \item \textit{Compare to sensible baselines.} Rather than reporting absolute performance in isolation, studies should contextualize results relative to unimodal baselines (text-only or vision-only) to demonstrate the benefits of multimodal approaches, random and majority-class baselines to establish task difficulty, prior work on the same dataset when available, and performance on related HR languages to quantify the LR gap.
\end{itemize}

As shown above, evaluation challenges remain a major problem in LR multimodal research, but some steps have already been taken towards fixing this gap. Although standard metrics represent a great starting point for evaluation, they are designed for English and often miss what matters for LR languages and their cultural contexts.
Solving these challenges requires creating culturally-appropriate benchmarks, using multiple and diverse evaluation metrics, as well as involving language communities in the evaluation process.

\section{Conclusion and Future Work}
\label{sec_conclusion}

\noindent
\textbf{Conclusion.}
Our survey has provided a comprehensive analysis of LMM-based approaches for LR languages, comprising 117 studies across 96 languages. Vision-language combinations dominate the current research landscape (65\% of surveyed works), with an increasing trend toward incorporating video and speech in recent works. We observed a concentration of research in South Asian languages (including Hindi, Bengali, Malayalam), Southeast Asian languages (Vietnamese, Javanese, Malay), Middle Eastern languages (Persian, Arabic) and African languages (Hausa, Amharic), while 42 other languages appear in only one study each.

The landscape of LMMs for LR languages has shown remarkable progress across multiple dimensions, from data creation to fusion techniques and architectural innovations. Projects like HVG, SEACrowd, and BVG highlight growing attention to creating high-quality multimodal resources for understudied languages. Recent successes with models such as Qalam, LaVy, and Amharic LLaVA \citep{2024-andersland} demonstrate that carefully designed multimodal strategies can effectively leverage limited resources, while adapting large-scale architectures for low-resource contexts.

\noindent
\textbf{Challenges and gaps.}
Our analysis reveals several critical challenges in the current landscape of LMMs for LR languages. A significant modality imbalance exists, with text-image pairs dominating research (65\% of studies), while audio and video modalities remain underexplored. This gap is particularly problematic for languages with strong oral traditions, where speech, tone and gesture carry essential linguistic information, with only 32\% of studies incorporating audio, despite its crucial importance for predominantly oral languages, and only 8.5\% of studies incorporating a video modality. We also identified persistent dataset scarcity and uneven language representation, with just three languages (Hindi, Arabic, Bengali) accounting for a disproportionate share of research attention. Technical limitations further constrain progress, as computational constraints limit the application of advanced fusion techniques in resource-constrained environments typical for LR contexts. Current cross-modal transfer methods struggle with catastrophic forgetting and inefficient knowledge transfer, particularly for languages that are structurally distant from high-resource counterparts. The field also lacks standardized evaluation frameworks for meaningful comparison across approaches, while recent work by Shen et al.~\cite{2024-shen} highlights significant safety challenges when deploying LLMs in multilingual contexts. Finally, sociolinguistic dimensions remain underexplored, including cultural representation, algorithmic bias, and potential impacts on language endangerment and revitalization efforts. These concerns are particularly acute given power imbalances between communities speaking low-resource languages and the primarily Western institutions developing these technologies.

Our study identifies three mechanisms through which LMMs may perpetuate digital inequalities. First, language model training inherently favors languages with larger training representation~\cite{navigli2023biases,wu2025inequalities}, introducing a bias towards modeling HR languages. Second, benchmarks derived from English (e.g.~translated MMLU) embed Western cultural assumptions that disadvantage LR language speakers even when linguistic accuracy is achieved~\cite{singh2024globalmmu}, introducing cultural biases in the evaluation. Third, computational requirements exclude researchers in LR language regions from model development, creating dependency on external institutions and biasing resource access. Addressing these biases requires community-centered approaches that prioritize local capacity building, alongside technical performance metrics.

\noindent\textbf{Future work.}
Based on the challenges identified above, we propose several key directions for future research.

For short-term development, we propose the following actionable research directions for benchmark and dataset creation: (1) extend Visual Genome-style multimodal datasets to at least 20 additional LR languages, prioritizing the 42 languages currently represented by only a single study; (2) develop speech-image paired corpora for tonal languages (e.g.~Yoruba, Igbo, Fongbe), where audio modality carries critical semantic distinctions absent in text; and (3) establish a standardized ``LR-MMBench'' evaluation suite with culturally-adapted visual question answering tasks, following SEACrowd's multilingual methodology, but incorporating non-Western visual contexts and evaluation protocols validated by native speakers.

To develop and improve LMMs for LR language, several concrete directions emerge from our analysis: (1) develop catastrophic forgetting mitigation strategies maintaining over 95\% source-language performance, while achieving over 80\% target-language performance for language pairs with fewer than 1,000 parallel sentences; (2) create language-agnostic visual encoders pre-trained on culturally-diverse image collections sourced from non-Western contexts, reducing the documented Western bias in current visual representations; and (3) establish explicit source-language selection guidelines based on typological similarity metrics (syntactic distance, shared writing systems, WALS features) to maximize positive transfer for specific target languages.

Several other research gaps require attention in future. Regarding the observed modality imbalance, researchers should prioritize incorporating audio and video for LR languages with limited writing traditions, enabling more robust applications that better reflect natural communication patterns, particularly for tonal languages and those where non-verbal communication is significant. For resource development, future work should advance synthetic data generation techniques (building on HVG, ELAICHI, Vintern-1B) and improve cross-lingual transfer methodologies (extending XtremeCLIP, LowCLIP) to accommodate greater linguistic diversity, while minimizing catastrophic forgetting. To overcome limitations in resource-constrained settings, researchers should investigate efficient fusion approaches including stacking-based late fusion, tensor fusion for complex interactions, and graphical fusion leveraging graph-based representations, all adapted for computational efficiency. Advancing adaptive integration through mechanisms that dynamically adjust the contribution of each modality based on input quality and task requirements will be crucial. Building on MRF \citep{2022-chauhan}, future work should explore hybrid approaches that combine strengths of different fusion strategies, while maintaining computational efficiency. Finally, adopting community-centered design approaches that address sociolinguistic dimensions alongside technical advances will ensure that developments benefit the intended language communities themselves.

Finally, we advocate for mandatory community engagement through: (i) participatory design frameworks requiring documented language community involvement in dataset creation, with explicit data governance and benefit-sharing agreements; (ii) open-source, mobile-first data collection libraries suitable for field conditions, where many LR languages are spoken; and (3) standardized model cards for LR multimodal systems, documenting limitations, cultural biases, and appropriate use cases, ensuring transparent communication with end-user communities.

\section*{Acknowledgments}

This research is supported by the project ``Romanian Hub for Artificial Intelligence - HRIA'', Smart Growth, Digitization and Financial Instruments Program, 2021-2027, MySMIS no.~351416. The authors thank reviewers for the constructive feedback.

\bibliographystyle{elsarticle-num}
\bibliography{refs}

@inproceedings{2018-dutta,
     title = "Multimodal Neural Machine Translation for Low-resource Language Pairs using Synthetic Data",
    author = "Dutta Chowdhury, Koel  and
      Hasanuzzaman, Mohammed  and
      Liu, Qun",
   
    booktitle = "Proceedings of the Workshop on Deep Learning Approaches for Low-Resource {NLP} (DeepLo)",
    year = "2018",
    url = "https://aclanthology.org/W18-3405/",
    doi = "10.18653/v1/W18-3405",
    pages = "33--42"
}

@inproceedings{10.5555/3540261.3540277,
author = {Tsimpoukelli, Maria and Menick, Jacob and Cabi, Serkan and Eslami, S. M. Ali and Vinyals, Oriol and Hill, Felix},
title = {Multimodal few-shot learning with frozen language models},
year = {2021},
booktitle = {Proceedings of the 35th International Conference on Neural Information Processing Systems (NeurIPS)},
pages={200--212},
url="https://proceedings.neurips.cc/paper/2021/file/01b7575c38dac42f3cfb7d500438b875-Paper.pdf"
}

@inproceedings{NEURIPS2023_e425b75b,
   title={Language is not all you need: aligning perception with language models},
  author={Huang, Shaohan and Dong, Li and Wang, Wenhui and Hao, Yaru and Singhal, Saksham and Ma, Shuming and Lv, Tengchao and Cui, Lei and Mohammed, Owais Khan and Patra, Barun and others},
  booktitle={Proceedings of the 37th International Conference on Neural Information Processing Systems (NeuIPS)},
  pages={72096--72109},
  year={2023},
url={https://dl.acm.org/doi/10.5555/3666122.3669277}
}

@inproceedings{2023-Rafailov-NeurIPS,
 author = {Rafailov, Rafael and Sharma, Archit and Mitchell, Eric and Manning, Christopher D. and Ermon, Stefano and Finn, Chelsea},
 booktitle = {Proceedings of the 37th International Conference on Neural Information Processing Systems (NeurIPS)},
 pages = {53728--53741},
 url ={https://proceedings.neurips.cc/paper_files/paper/2023/file/a85b405ed65c6477a4fe8302b5e06ce7-Paper-Conference.pdf},
 volume = {36},
 year = {2023},
title="{Direct Preference Optimization:
Your Language Model is Secretly a Reward Model}"
}

@article{lent-etal-2024-creoleval,
    title = "{C}reole{V}al: Multilingual Multitask Benchmarks for Creoles",
    author = {Lent, Heather  and
      Tatariya, Kushal  and
      Dabre, Raj  and
      Chen, Yiyi  and
      Fekete, Marcell  and
      Ploeger, Esther  and
      Zhou, Li  and
      Armstrong, Ruth-Ann  and
      Eijansantos, Abee  and
      Malau, Catriona  and
      others},
    journal = "Transactions of the Association for Computational Linguistics",
    volume = "12",
    year = "2024",
    url = "https://aclanthology.org/2024.tacl-1.53/",
    doi = "10.1162/tacl_a_00682",
    pages = "950--978"
}

@inproceedings{driess2023palm,
  author = {Driess, Danny and Xia, Fei and Sajjadi, Mehdi S. M. and Lynch, Corey and Chowdhery, Aakanksha and Ichter, Brian and Wahid, Ayzaan and Tompson, Jonathan and Vuong, Quan and Yu, Tianhe and others},
title = "{PaLM-E: an embodied multimodal language model}",
year = {2023},
booktitle = {Proceedings of the 40th International Conference on Machine Learning (ICML)},
pages = {8469--8488},
numpages = {20},
url={https://dl.acm.org/doi/10.5555/3618408.3618748}
}

@article{2018-najadat,
    title="{Multimodal sentiment analysis of Arabic videos}",
  author={Najadat, Hassan and Abushaqra, Ftoon},
  journal={Journal of Image and Graphics},
  volume={6},
  number={1},
  pages={39--43},
  year={2018},
url={https://www.joig.net/index.php?m=content&c=index&a=show&catid=47&id=173}
}

@article{2019-parida,
    title="{Hindi visual genome: A dataset for multi-modal English to Hindi machine translation}",
  author={Parida, Shantipriya and Bojar, Ond{\v{r}}ej and Dash, Satya Ranjan},
  journal={Computaci{\'o}n y Sistemas},
  volume={23},
  number={4},
  pages={1499--1505},
  year={2019},
doi={10.13053/cys-23-4-3294},
url={https://www.cys.cic.ipn.mx/ojs/index.php/CyS/article/view/3294}
}

@inproceedings{2019-chakravarthi,
        title = "Multilingual Multimodal Machine Translation for {D}ravidian Languages utilizing Phonetic Transcription",
    author = "Chakravarthi, Bharathi Raja  and
      Priyadharshini, Ruba  and
      Stearns, Bernardo  and
      Jayapal, Arun  and
      Sridevy, S  and
      Arcan, Mihael  and
      Zarrouk, Manel  and
      McCrae, John P",
    
    booktitle = "Proceedings of the 2nd Workshop on Technologies for MT of Low Resource Languages (LoResMT)",
    year = "2019",
    url = "https://aclanthology.org/W19-6809/",
    pages = "56--63"
}

@inproceedings{2019-wu,
    author = {Wu, Yike and Zhao, Shiwan and Chen, Jia and Zhang, Ying and Yuan, Xiaojie and Su, Zhong},
booktitle = {Proceedings of IEEE International Conference on Multimedia and Expo (ICME)},
title = {Improving Captioning for Low-Resource Languages by Cycle Consistency},
year = {2019},
pages = {362--367},
doi = {10.1109/ICME.2019.00070},
url = {https://ieeexplore.ieee.org/document/8784910},
}

@inproceedings{2020-laskar,
     title = "Multimodal Neural Machine Translation for {E}nglish to {H}indi",
    author = "Laskar, Sahinur Rahman  and
      Khilji, Abdullah Faiz Ur Rahman  and
      Pakray, Partha  and
      Bandyopadhyay, Sivaji",
   
    booktitle = "Proceedings of the 7th Workshop on Asian Translation (WAT)",
    year = "2020",
    url = "https://aclanthology.org/2020.wat-1.11/",
    doi = "10.18653/v1/2020.wat-1.11",
    pages = "109--113",
}

@article{2020-azani,
    author={Al-Azani, Sadam and El-Alfy, El-Sayed M.},
  journal={IEEE Access}, 
  title={Enhanced Video Analytics for Sentiment Analysis Based on Fusing Textual, Auditory and Visual Information}, 
  year={2020},
  volume={8},
  number={},
  pages={136843--136857},
  doi={10.1109/ACCESS.2020.3011977},
  url={https://ieeexplore.ieee.org/document/9148603}
}

@inproceedings{2021-laskar-english-assamese,
    author={Laskar, Sahinur Rahman and Paul, Bishwaraj and Paudwal, Siddharth and Gautam, Pranjit and Biswas, Nirmita and Pakray, Partha},
  booktitle={Proceedings of International Conference on Computational Performance Evaluation (ComPE)}, 
  title="{Multimodal Neural Machine Translation for English–Assamese Pair}", 
  year={2021},
  pages={387--392},
  doi={10.1109/ComPE53109.2021.9752181},
  url={https://ieeexplore.ieee.org/document/9752181}
}

@inproceedings{2023-laskar,
title = "{English-Assamese Multimodal Neural Machine Translation using Transliteration-based Phrase Augmentation Approach}",
booktitle = {Proceedings of International conference on Machine Learning and Data Engineering (ICMLDE)},
volume = {218},
pages = {979--988},
year = {2023},
issn = {1877-0509},
doi = {10.1016/j.procs.2023.01.078},
url = {https://www.sciencedirect.com/science/article/pii/S1877050923000789},
author = {Sahinur Rahman Laskar and Bishwaraj Paul and Partha Pakray and Sivaji Bandyopadhyay},
}

@article{2024-tayir,
author = {Tayir, Turghun and Li, Lin},
title = {Unsupervised Multimodal Machine Translation for Low-resource Distant Language Pairs},
year = {2024},
issue_date = {April 2024},
volume = {23},
number = {4},
issn = {2375-4699},
url = {https://dl.acm.org/doi/10.1145/3652161},
doi = {10.1145/3652161},
journal = {ACM Transactions on Asian and Low-Resource Language Information Processing},
pages = {1--22},
}

@inproceedings{2021-laskar-hindi,
title = "Improved {E}nglish to {H}indi Multimodal Neural Machine Translation",
    author = "Laskar, Sahinur Rahman  and
      Khilji, Abdullah Faiz Ur Rahman  and
      Kaushik, Darsh  and
      Pakray, Partha  and
      Bandyopadhyay, Sivaji",
    
    booktitle = "Proceedings of the 8th Workshop on Asian Translation (WAT)",
    year = "2021",
    url = "https://aclanthology.org/2021.wat-1.17/",
    doi = "10.18653/v1/2021.wat-1.17",
    pages = "155--160"
}

@inproceedings{youcef2021arabic,
  author={Youcef, Fatima Zohra and Barigou, Fatiha},
  booktitle={Proceedings of 22nd International Arab Conference on Information Technology (ACIT)}, 
  title={Arabic language investigation in the context of unimodal and multimodal sentiment analysis}, 
  year={2021},
  pages={1--7},
  doi={10.1109/ACIT53391.2021.9677274},
  url={https://ieeexplore.ieee.org/document/9677274}
}

@article{2021-dashtipour,
   author = {Dashtipour, Kia and Gogate, Mandar and Cambria, Erik and Hussain, Amir},
title = "{A novel context-aware multimodal framework for Persian sentiment analysis}",
year = {2021},
volume = {457},
number = {C},
issn = {0925-2312},
url = {https://www.sciencedirect.com/science/article/abs/pii/S0925231221002666},
doi = {10.1016/j.neucom.2021.02.020},
journal = {Neurocomputing},
pages = {377--388},
numpages = {12},
}

@inproceedings{2021-sanayai,
    title = "An Experiment on Speech-to-Text Translation Systems for {M}anipuri to {E}nglish on Low Resource Setting",
    author = "Sanayai Meetei, Loitongbam  and
      Rahul, Laishram  and
      Singh, Alok  and
      Singh, Salam Michael  and
      Singh, Thoudam Doren  and
      Bandyopadhyay, Sivaji",
   
    booktitle = "Proceedings of the 18th International Conference on Natural Language Processing (ICON)",
    year = "2021",
    url = "https://aclanthology.org/2021.icon-main.8/",
    pages = "54--63",
}

@inproceedings{2021-parida,
     title = "Multimodal Neural Machine Translation System for {E}nglish to {B}engali",
    author = "Parida, Shantipriya  and
      Panda, Subhadarshi  and
      Biswal, Satya Prakash  and
      Kotwal, Ketan  and
      Sen, Arghyadeep  and
      Dash, Satya Ranjan  and
      Motlicek, Petr",
  
    booktitle = "Proceedings of the First Workshop on Multimodal Machine Translation for Low Resource Languages (MMTLRL)",
    year = "2021",
    url = "https://aclanthology.org/2021.mmtlrl-1.6/",
    pages = "31--39",
}

@inproceedings{2021-jain,
    title = "{MURAL}: Multimodal, Multitask Representations Across Languages",
    author = "Jain, Aashi  and
      Guo, Mandy  and
      Srinivasan, Krishna  and
      Chen, Ting  and
      Kudugunta, Sneha  and
      Jia, Chao  and
      Yang, Yinfei  and
      Baldridge, Jason",
   
    booktitle = "Findings of the Association for Computational Linguistics: 
Empirical Methods in Natural Language Processing (EMNLP)",
    year = "2021",
    url = "https://aclanthology.org/2021.findings-emnlp.293/",
    doi = "10.18653/v1/2021.findings-emnlp.293",
    pages = "3449--3463"
}

@article{2021-sehar,
    author={Sehar, Urooba and Kanwal, Summrina and Dashtipur, Kia and Mir, Usama and Abbasi, Ubaid and Khan, Faiza},
  journal={IEEE Access}, 
  title={Urdu Sentiment Analysis via Multimodal Data Mining Based on Deep Learning Algorithms}, 
  year={2021},
  volume={9},
  number={},
  pages={153072--153082},
  doi={10.1109/ACCESS.2021.3122025},
  url={https://ieeexplore.ieee.org/document/9583225}
}

@inproceedings{2022-laskar,
      title = "Investigation of {E}nglish to {H}indi Multimodal Neural Machine Translation using Transliteration-based Phrase Pairs Augmentation",
    author = "Laskar, Sahinur Rahman  and
      Singh, Rahul  and
      Karim, Md Faizal  and
      Manna, Riyanka  and
      Pakray, Partha  and
      Bandyopadhyay, Sivaji",
    booktitle = "Proceedings of the 9th Workshop on Asian Translation (WAT)",
    year = "2022",
    url = "https://aclanthology.org/2022.wat-1.15/",
    pages = "117--122"
}

@article{2022-shi,
   title={Adding Visual Information to Improve Multimodal Machine Translation for Low-Resource Language},
  author={Shi, Xiayang and Yu, Zhenqiang},
  journal={Mathematical Problems in Engineering},
  volume={2022},
  number={1},
  pages={5483535},
  year={2022},
doi={10.1155/2022/5483535},
url={https://onlinelibrary.wiley.com/doi/10.1155/2022/5483535}
}

@article{wu2022pairs,
      author = {Wu, Yike and Zhao, Shiwan and Zhang, Ying and Yuan, Xiaojie and Su, Zhong},
title = {When Pairs Meet Triplets: Improving Low-Resource Captioning via Multi-Objective Optimization},
year = {2022},
volume = {18},
number = {3},
pages={1--20},
issn = {1551-6857},
url = {https://dl.acm.org/doi/10.1145/3492325},
doi = {10.1145/3492325},
journal = {ACM Transactions on Multimedia Computing, Communications, and Applications},
numpages = {20},
}

@inproceedings{2022-abdulmumin,
        title = "{H}ausa Visual Genome: A Dataset for Multi-Modal {E}nglish to {H}ausa Machine Translation",
    author = "Abdulmumin, Idris  and
      Dash, Satya Ranjan  and
      Dawud, Musa Abdullahi  and
      Parida, Shantipriya  and
      Muhammad, Shamsuddeen  and
      Ahmad, Ibrahim Sa{'}id  and
      Panda, Subhadarshi  and
      Bojar, Ond{\v{r}}ej  and
      Galadanci, Bashir Shehu  and
      Bello, Bello Shehu",
  
    booktitle = "Proceedings of the Thirteenth Language Resources and Evaluation Conference (LREC)",
    year = "2022",
    url = "https://aclanthology.org/2022.lrec-1.694/",
    pages = "6471--6479"

}

@inproceedings{2023-chen,
       author={Chen, Yutong and Wei, Fangyun and Sun, Xiao and Wu, Zhirong and Lin, Stephen},
  booktitle={Proceedings of IEEE/CVF Conference on Computer Vision and Pattern Recognition (CVPR)}, 
  title={A Simple Multi-Modality Transfer Learning Baseline for Sign Language Translation}, 
  year={2022},
  volume={},
  number={},
  pages={5110--5120},
  doi={10.1109/CVPR52688.2022.00506},
  url={https://ieeexplore.ieee.org/document/9879103/}}

@inproceedings{2022-chauhan,
author = {Chauhan, Dushyant Singh and Ekbal, Asif and Bhattacharyya, Pushpak},
title = {An Efficient Fusion Mechanism for Multimodal Low-resource Setting},
year = {2022},
isbn = {9781450387323},
url = {https://doi.org/10.1145/3477495.3531900},
doi = {10.1145/3477495.3531900},
booktitle = {Proceedings of the 45th International ACM SIGIR Conference on Research and Development in Information Retrieval (SIGIR)},
pages = {2583--2588},
numpages = {6},
}

@article{2024-qu,
  title={Mitigating Multilingual Hallucination in Large Vision-Language Models},
  author={Xiaoye Qu and Mingyang Song and Wei Wei and Jianfeng Dong and Yu Cheng},
journal={arXiv preprint arXiv:2408.00550},  
year={2024},
  url={https://arxiv.org/abs/2408.00550}
}

@inproceedings{farsi-etal-2025-persian,
    title = "{P}ersian in a Court: Benchmarking {VLM}s In {P}ersian Multi-Modal Tasks",
    author = "Farsi, Farhan  and
      Shariati Motlagh, Shahriar  and
      Bali, Shayan  and
      Sabouri, Sadra  and
      Momtazi, Saeedeh",
   
    booktitle = "Proceedings of the First Workshop of Evaluation of Multi-Modal Generation (EvalMG)",
    year = "2025",
    url = "https://aclanthology.org/2025.evalmg-1.5/",
    pages = "52--56"
}

@inproceedings{2022-sen,
  title="{Bengali Visual Genome: A multimodal dataset for machine translation and image captioning}",
  author={Sen, Arghyadeep and Parida, Shantipriya and Kotwal, Ketan and Panda, Subhadarshi and Bojar, Ond{\v{r}}ej and Dash, Satya Ranjan},
  booktitle={Proceedings of the 9th International Conference on Frontiers in Intelligent Computing: Theory and Applications (FICTA)},
  pages={63--70},
  year={2021},
doi={10.1007/978-981-16-6624-7_7},
url = {https://link.springer.com/chapter/10.1007/978-981-16-6624-7_7}
}

@inproceedings{parida-etal-2025-ovqa,
    title = "{{OVQA}: A Dataset for Visual Question Answering and Multimodal Research in {O}dia Language}",
    author = "Parida, Shantipriya  and
      Sahoo, Shashikanta  and
      Sekhar, Sambit  and
      Sahoo, Kalyanamalini  and
      Kotwal, Ketan  and
      Khosla, Sonal  and
      Dash, Satya Ranjan  and
      Bose, Aneesh  and
      Kohli, Guneet Singh  and
      Lenka, Smruti Smita  and
      Bojar, Ond{\v{r}}ej",
    
    booktitle = "Proceedings of the First Workshop on Natural Language Processing for Indo-Aryan and Dravidian Languages (IndoNLP)",
    year = "2025",
    url = "https://aclanthology.org/2025.indonlp-1.7/",
    pages = "58--66",
}

@inproceedings{2022-h-o,
   author={Lekshmy, H.O. and Jayaraman, Swaminathan},
  booktitle={Proceedings of 6th International Conference on Intelligent Computing and Control Systems (ICICCS)}, 
  title="{English-Malayalam Vision aid with Multi Modal Machine Learning Technologies}", 
  year={2022},
  pages={1469--1476},
  doi={10.1109/ICICCS53718.2022.9788187},
  url={https://ieeexplore.ieee.org/document/9788187}}

@article{2022-mamta,
author = {Mamta and Ekbal, Asif and Bhattacharyya, Pushpak},
title = {Exploring Multi-lingual, Multi-task, and Adversarial Learning for Low-resource Sentiment Analysis},
year = {2022},
issue_date = {September 2022},
volume = {21},
number = {5},
issn = {2375-4699},
url = {https://dl.acm.org/doi/10.1145/3514498},
doi = {10.1145/3514498},
journal = {ACM Transactions on Asian and Low-Resource Language Information Processing},
pages = {104},
numpages = {19},
}

@inproceedings{2022-debele,
    author={Debele, Abreham Gebremedin and Woldeyohannis, Michael Melese},
  booktitle={Proceedings of International Conference on Information and Communication Technology for Development for Africa (ICT4DA)}, 
  title={Multimodal {A}mharic Hate Speech Detection Using Deep Learning}, 
  year={2022},
  pages={102--107},
  doi={10.1109/ICT4DA56482.2022.9971436},
  url={https://ieeexplore.ieee.org/document/9971436/}
}

@inproceedings{2022-karim,
  title={Multimodal hate speech detection from {B}engali memes and texts},
  author={Karim, Md Rezaul and Dey, Sumon Kanti and Islam, Tanhim and Shajalal, Md and Chakravarthi, Bharathi Raja},
  booktitle={Proceedings of the International Conference on Speech and Language Technologies for Low-Resource Languages (SPELLL)},
  pages={293--308},
  year={2022},
url = {https://link.springer.com/chapter/10.1007/978-3-031-33231-9_21}
}

@inproceedings{2022-hossain-mute,
 title = "{MUTE}: A Multimodal Dataset for Detecting Hateful Memes",
    author = "Hossain, Eftekhar  and
      Sharif, Omar  and
      Hoque, Mohammed Moshiul",
   
    booktitle = "Proceedings of the 2nd Conference of the Asia-Pacific Chapter of the Association for Computational Linguistics and the 12th International Joint Conference on Natural Language Processing: Student Research Workshop (AACL-IJCNLP)",
    year = "2022",
    url = "https://aclanthology.org/2022.aacl-srw.5/",
    doi = "10.18653/v1/2022.aacl-srw.5",
    pages = "32--39",
}

@inproceedings{2023-dash,
        title = "{BITS}-{P} at {WAT} 2023: Improving {I}ndic Language Multimodal Translation by Image Augmentation using Diffusion Models",
    author = "Dash, Amulya  and
      Gupta, Hrithik Raj  and
      Sharma, Yashvardhan",
   
    booktitle = "Proceedings of the 10th Workshop on Asian Translation (WAT)",
    year = "2023",
    url = "https://aclanthology.org/2023.wat-1.3/",
    pages = "41--45"
}

@inproceedings{2023-haouhat,
  title={Towards {A}rabic multimodal dataset for sentiment analysis},
  author={Haouhat, Abdelhamid and Bellaouar, Slimane and Nehar, Attia and Cherroun, Hadda},
  booktitle={Proceedings of Fourth International Conference on Intelligent Data Science Technologies and Applications (IDSTA)},
  pages={126--133},
  year={2023},
doi = {10.1109/IDSTA58916.2023.10317847},
url={https://ieeexplore.ieee.org/document/10317847}
}

@inproceedings{2023-sikasote,
      title = "{BIG}-{C}: a Multimodal Multi-Purpose Dataset for {B}emba",
    author = "Sikasote, Claytone  and
      Mukonde, Eunice  and
      Alam, Md Mahfuz Ibn  and
      Anastasopoulos, Antonios",

    booktitle = "Proceedings of the 61st Annual Meeting of the Association for Computational Linguistics (ACL)",
    year = "2023",
    url = "https://aclanthology.org/2023.acl-long.115/",
    doi = "10.18653/v1/2023.acl-long.115",
    pages = "2062--2078",
}

@inproceedings{2023-alalem,
   author={Alalem, Sohaila and Zaghloul, Mohamed Saad and Badawy, Osama},
  booktitle={Proceedings of 24th International Arab Conference on Information Technology (ACIT)}, 
  title="{A Novel Deep Learning Multi-Modal Sentiment Analysis Model for English and Egyptian Arabic Dialects Using Audio and Text}", 
  year={2023},
  pages={1--5},
  doi={10.1109/ACIT58888.2023.10453875},
  url={https://ieeexplore.ieee.org/document/10453875}
}

@inproceedings{2023-wang-adapting,
  author={Wang, Ying and Pfeiffer, Jonas and Carion, Nicolas and LeCun, Yann and Kamath, Aishwarya},
  booktitle={Proceedings of IEEE/CVF Conference on Computer Vision and Pattern Recognition Workshops (CVPRW)}, 
  title={Adapting Grounded Visual Question Answering Models to Low Resource Languages}, 
  year={2023},
  pages={2596--2605},
  doi={10.1109/CVPRW59228.2023.00258},
url = {https://ieeexplore.ieee.org/document/10208296/}
}

@inproceedings{2022-hamzah,
       author={Luqman, Hamzah},
  booktitle={Proceedings of IEEE 17th International Conference on Automatic Face and Gesture Recognition (FG)}, 
  title="{ArabSign: A Multi-modality Dataset and Benchmark for Continuous Arabic Sign Language Recognition}", 
  year={2023},
  volume={},
  number={},
  pages={1--8},
  doi={10.1109/FG57933.2023.10042720},
url = {https://ieeexplore.ieee.org/document/10042720}
}

@inproceedings{2023-santos,
      title = "{CAPIVARA}: Cost-Efficient Approach for Improving Multilingual {CLIP} Performance on Low-Resource Languages",
    author = "dos Santos, Gabriel Oliveira  and
      Braga Moreira, Diego Alysson  and
      Ferreira, Alef Iury  and
      Silva, Jhessica  and
      Pereira, Luiz  and
      Bueno, Pedro  and
      Sousa, Thiago  and
      Maia, Helena  and
      Da Silva, N{\'a}dia  and
      Colombini, Esther  and
      Pedrini, Helio  and
      Avila, Sandra",
    booktitle = "Proceedings of the 3rd Workshop on Multi-lingual Representation Learning (MRL)",
    year = "2023",
    url = "https://aclanthology.org/2023.mrl-1.15/",
    doi = "10.18653/v1/2023.mrl-1.15",
    pages = "184--207"
}

@article{2023-meetei-cues,
title = {Do cues in a video help in handling rare words in a machine translation system under a low-resource setting?},
journal = {Natural Language Processing Journal},
volume = {3},
pages = {100016},
year = {2023},
issn = {2949-7191},
doi = {https://doi.org/10.1016/j.nlp.2023.100016},
url = {https://www.sciencedirect.com/science/article/pii/S2949719123000134},
author = {Loitongbam Sanayai Meetei and Alok Singh and Thoudam Doren Singh and Sivaji Bandyopadhyay},
}

@article{2023-meetei,
author = {Meetei, Loitongbam and Singh, Thoudam Doren and Bandyopadhyay, Sivaji},
year = {2024},
pages = {13137–13157},
title = {Exploiting multiple correlated modalities can enhance low-resource machine translation quality},
volume = {83},
journal = {Multimedia Tools and Applications},
doi = {10.1007/s11042-023-15721-2},
url = {https://link.springer.com/article/10.1007/s11042-023-15721-2}
}

@inproceedings{2023-saichyshyna,
        title = "Extension {M}ulti30{K}: Multimodal Dataset for Integrated Vision and Language Research in {U}krainian",
    author = "Saichyshyna, Nataliia  and
      Maksymenko, Daniil  and
      Turuta, Oleksii  and
      Yerokhin, Andriy  and
      Babii, Andrii  and
      Turuta, Olena",
    booktitle = "Proceedings of the Second Ukrainian Natural Language Processing Workshop (UNLP)",
    year = "2023",
    url = "https://aclanthology.org/2023.unlp-1.7/",
    doi = "10.18653/v1/2023.unlp-1.7",
    pages = "54--61",
}

@inproceedings{2023-premjith,
    title = "Findings of the Shared Task on Multimodal Abusive Language Detection and Sentiment Analysis in {T}amil and {M}alayalam",
    author = "Premjith, B.  and
      Jyothish Lal, G.  and
      Sowmya, V.  and
      Chakravarthi, Bharathi Raja  and
      Natarajan, Rajeswari  and
      Nandhini, K.  and
      Murugappan, Abirami  and
      Bharathi, B.  and
      Kaushik, M.  and
      Prasanth, Sn  and
      Aswin Raj, R.  and
      Vijai Simmon, S.",
   
    booktitle = "Proceedings of the Third Workshop on Speech and Language Technologies for Dravidian Languages (DravidianLangTech)",
    year = "2023",
    url = "https://aclanthology.org/2023.dravidianlangtech-1.10/",
    pages = "72--79"
}

@inproceedings{2023-meetei-hindi,
  title="{Hindi to English Multimodal Machine Translation on News Dataset in Low Resource Setting}",
  author={Loitongbam Sanayai Meetei and Salam Michael Singh and Alok Singh and Ringki Das and Thoudam Doren Singh and Sivaji Bandyopadhyay},
  booktitle={Proceedings of International Conference on Machine Learning and Data Engineering (ICMLDE)},
  year={2023},
  doi = {10.1016/j.procs.2023.01.186},
  url = {https://www.sciencedirect.com/science/article/pii/S1877050923001862},
  pages={2102--2109},
  volume={218}

}

@inproceedings{2023-kim,
  title={Lip reading for low-resource languages by learning and combining general speech knowledge and language-specific knowledge},
  author={Kim, Minsu and Yeo, Jeong Hun and Choi, Jeongsoo and Ro, Yong Man},
  booktitle={Proceedings of the IEEE/CVF International Conference on Computer Vision (ICCV)},
  pages={15359--15371},
  year={2023},
  doi = {10.1109/ICCV51070.2023.01409},
  url = {https://ieeexplore.ieee.org/document/10377080/}
}

@inproceedings{2023-anwar,
      author = {Anwar, Mohamed and Shi, Bowen and Goswami, Vedanuj and Hsu, Wei-Ning and Pino, Juan and Wang, Changhan},
year = {2023},
pages = {4064--4068},
booktitle ={Proceedings of Conference of the International Speech Communication Association (INTERSPEECH)},
title = "{MuAViC: A Multilingual Audio-Visual Corpus for Robust Speech Recognition and Robust Speech-to-Text Translation}",
doi = {10.21437/Interspeech.2023-2279},
url = {https://www.isca-archive.org/interspeech_2023/anwar23_interspeech.html}
}

@article{2024-roken,
author = {Al Roken, Noora and Barlas, Gerassimos},
title = "{Multimodal Arabic emotion recognition using deep learning}",
year = {2023},
volume = {155},
number = {C},
url = {https://doi.org/10.1016/j.specom.2023.103005},
doi = {10.1016/j.specom.2023.103005},
journal = {Speech Communication},
pages = {103005},
}

@article{2023-albalawi,
  author={Albalawi, Rasha M. and Jamal, Amani T. and Khadidos, Alaa O. and Alhothali, Areej M.},
  journal={IEEE Access}, 
  title={Multimodal {A}rabic Rumors Detection}, 
  year={2023},
  volume={11},
  number={},
  pages={9716--9730},
  doi={10.1109/ACCESS.2023.3240373},
  url={https://ieeexplore.ieee.org/document/10026837}
}

@inproceedings{2023-tang-xtremeclip,
    title = "{X}treme{CLIP}: Extremely Parameter-efficient Tuning for Low-resource Vision Language Understanding",
    author = "Tang, Moming  and
      Wang, Chengyu  and
      Wang, Jianing  and
      Tan, Chuanqi  and
      Huang, Songfang  and
      Chen, Cen  and
      Qian, Weining",
  
    booktitle = "Findings of the Association for Computational Linguistics (ACL)",
    year = "2023",
    url = "https://aclanthology.org/2023.findings-acl.397/",
    doi = "10.18653/v1/2023.findings-acl.397",
    pages = "6368--6376"
}

@inproceedings{2023-parida,
     title = "{H}a{VQA}: A Dataset for Visual Question Answering and Multimodal Research in {H}ausa Language",
    author = "Parida, Shantipriya  and
      Abdulmumin, Idris  and
      Muhammad, Shamsuddeen Hassan  and
      Bose, Aneesh  and
      Kohli, Guneet Singh  and
      Ahmad, Ibrahim Said  and
      Kotwal, Ketan  and
      Deb Sarkar, Sayan  and
      Bojar, Ond{\v{r}}ej  and
      Kakudi, Habeebah",
    booktitle = "Findings of the Association for Computational Linguistics (ACL)",
    year = "2023",
    url = "https://aclanthology.org/2023.findings-acl.646/",
    doi = "10.18653/v1/2023.findings-acl.646",
    pages = "10162--10183",
}

@inproceedings{jin2022good,
    title = "A Good Prompt Is Worth Millions of Parameters: Low-resource Prompt-based Learning for Vision-Language Models",
    author = "Jin, Woojeong  and
      Cheng, Yu  and
      Shen, Yelong  and
      Chen, Weizhu  and
      Ren, Xiang",
   
    booktitle = "Proceedings of the 60th Annual Meeting of the Association for Computational Linguistics (ACL)",
    year = "2022",
    url = "https://aclanthology.org/2022.acl-long.197/",
    doi = "10.18653/v1/2022.acl-long.197",
    pages = "2763--2775"
}

@inproceedings{2022-hossain,
     title = "{M}emo{S}en: A Multimodal Dataset for Sentiment Analysis of Memes",
    author = "Hossain, Eftekhar  and
      Sharif, Omar  and
      Hoque, Mohammed Moshiul",
    
    booktitle = "Proceedings of the Thirteenth Language Resources and Evaluation Conference (LREC)",
    year = "2022",
    url = "https://aclanthology.org/2022.lrec-1.165/",
    pages = "1542--1554"
}

@inproceedings{2023-elahi,
        title={Explainable multimodal sentiment analysis on {B}engali memes},
  author={Elahi, Kazi Toufique and Rahman, Tasnuva Binte and Shahriar, Shakil and Sarker, Samir and Joy, Sajib Kumar Saha and Shah, Faisal Muhammad},
  booktitle={Proceedings of 26th International Conference on Computer and Information Technology (ICCIT)},
  pages={1--6},
  year={2023},
  doi = {10.1109/ICCIT60459.2023.10441342},
  url = {https://ieeexplore.ieee.org/document/10441342}
}

@article{2024-taylor,
author = {Taylor, Serena and Fauzi, Fariza},
title = "{Multimodal Sentiment Analysis for the Malay Language: New Corpus using CNN-based Framework}",
year = {2024},
issn = {2375-4699},
url = {https://dl.acm.org/doi/10.1145/3703445},
doi = {10.1145/3703445},
journal = {ACM Transactions on Asian and Low-Resource Language Information Processing},
volume = {24},
pages={1--26}
}

@inproceedings{2024-hatami,
    title = "{E}nglish-to-Low-Resource Translation: A Multimodal Approach for {H}indi, {M}alayalam, {B}engali, and {H}ausa",
    author = "Hatami, Ali  and
      Banerjee, Shubhanker  and
      Arcan, Mihael  and
      Buitelaar, Paul  and
      Philip McCrae, John",
   
    booktitle = "Proceedings of the Ninth Conference on Machine Translation (WMT)",
    year = "2024",
    url = "https://aclanthology.org/2024.wmt-1.76/",
    doi = "10.18653/v1/2024.wmt-1.76",
    pages = "815--822"
}

@inproceedings{2024-haq,
    title = "{DCU} {ADAPT} at {WMT}24: {E}nglish to Low-resource Multi-Modal Translation Task",
    author = "Haq, Sami  and
      Huidrom, Rudali  and
      Castilho, Sheila",
  
    booktitle = "Proceedings of the Ninth Conference on Machine Translation (WMT)",
    year = "2024",
    url = "https://aclanthology.org/2024.wmt-1.75/",
    doi = "10.18653/v1/2024.wmt-1.75",
    pages = "810--814",
}

@article{2024-amalas,
        title={A multilingual training strategy for low resource Text to Speech},
  author={Asma Amalas and Mounir Ghogho and Mohamed Chetouani and Rachid Oulad Haj Thami},
journal={arXiv preprint arXiv:2409.01217},
year={2024},
  url={https://arxiv.org/abs/2409.01217}
}

@article{young2014image,
   title = "From image descriptions to visual denotations: New similarity metrics for semantic inference over event descriptions",
    author = "Young, Peter  and
      Lai, Alice  and
      Hodosh, Micah  and
      Hockenmaier, Julia",
    
    journal = "Transactions of the Association for Computational Linguistics",
    volume = "2",
    year = "2014",
    url = "https://aclanthology.org/Q14-1006/",
    doi = "10.1162/tacl_a_00166",
    pages = "67--78",
}

@article{2024-arifin,
author = {Arifin, Fatchul and Nasuha, Aris and Priambodo, Ardy and Winursito, Anggun and Gunawan, Teddy},
year = {2024},
title = "{Advanced Multimodal Emotion Recognition for Javanese Language Using Deep Learning}",
volume = {12},
number={3},
pages={503--515},
journal = {Indonesian Journal of Electrical Engineering and Informatics},
doi = {10.52549/ijeei.v12i3.5662},
url = {https://section.iaesonline.com/index.php/IJEEI/article/view/5662}
}

@article{2024-andersland,
      title="{Amharic LLaMA and LLaVA: Multimodal LLMs for Low Resource Languages}", 
      author={Michael Andersland},
      year={2024},
      journal={arXiv preprint arXiv:2403.06354},
      doi={10.48550/arXiv.2403.06354},
      url={https://arxiv.org/abs/2403.06354}, 
}

@article{2024-onuoha,
author = {Onuoha, C.E. and Uba, E.},
year = {2024},
title = "{An analysis of minimal pairs in Igbo using a multimodal approach to speech perception}",
journal = {Unizik Journal of Arts and Humanities},
volume={25},
  pages={31--50},
doi={10.4314/ujah.v25i1.2},
url={https://www.ajol.info/index.php/ujah/article/view/272304}

}

@inproceedings{2024-Alwajih-dallah,
       title = "Dallah: A Dialect-Aware Multimodal Large Language Model for {A}rabic",
    author = "Alwajih, Fakhraddin  and
      Bhatia, Gagan  and
      Abdul-Mageed, Muhammad",
    
    booktitle = "Proceedings of the Second Arabic Natural Language Processing Conference (ArabicNLP)",
    year = "2024",
    url = "https://aclanthology.org/2024.arabicnlp-1.27/",
    doi = "10.18653/v1/2024.arabicnlp-1.27",
    pages = "320--336"
}

@inproceedings{2024-jigar,
        title = "Detecting Hate Speech in {A}mharic Using Multimodal Analysis of Social Media Memes",
    author = "Jigar, Melese Ayichlie  and
      Ayele, Abinew Ali  and
      Yimam, Seid Muhie  and
      Biemann, Chris",
    
    booktitle = "Proceedings of the Fourth Workshop on Threat, Aggression {\&} Cyberbullying (TRAC)",
    year = "2024",
    url = "https://aclanthology.org/2024.trac-1.10/",
    pages = "85--95",
}

@article{2021-Chakravarthi,
      title="{DravidianMultiModality: A Dataset for Multi-modal Sentiment Analysis in Tamil and Malayalam}", 
      author={Chakravarthi, Bharathi Raja and {Parameswaran P.K.}, Jishnu  and {Premjith B} and Soman, K.P. and Ponnusamy, Rahul and Kumaresan, Prasanna Kumar and Thamburaj, Kingston Pal and McCrae, John P.},

        
      year={2021},
journal={arXiv preprint arXiv:2106.04853},
url={https://arxiv.org/abs/2106.04853}, 
}

@article{2024-tran-lavy,
  title="{LaVy: Vietnamese Multimodal Large Language Model}",
  author={Chi Tran and Huong L{\^e} Thanh},
journal={arXiv preprint arXiv:2404.07922},
year={2024},
  url={https://arxiv.org/abs/2404.07922}
}

@inproceedings{2024-alam,
    title = "{LLM}s for Low Resource Languages in Multilingual, Multimodal and Dialectal Settings",
    author = "Alam, Firoj  and
      Chowdhury, Shammur Absar  and
      Boughorbel, Sabri  and
      Hasanain, Maram",
   
    booktitle = "Proceedings of the 18th Conference of the European Chapter of the Association for Computational Linguistics: Tutorial Abstracts (EACL)",
    year = "2024",
    url = "https://aclanthology.org/2024.eacl-tutorials.5/",
    pages = "27--33"
}

@article{2023-Asgarov,
      title="{LowCLIP: Adapting the CLIP Model Architecture for Low-Resource Languages in Multimodal Image Retrieval Task}", 
      author={Asgarov, Ali and Rustamov, Samir},
  journal={arXiv preprint arXiv:2408.13909},
      year={2024},      
    doi = {10.48550/arXiv.2408.13909},
    url = {https://arxiv.org/abs/2408.13909}
}

@article{2023-gain,
  title="{Impact of Visual Context on Noisy Multimodal NMT: An Empirical Study for English to Indian Languages}",
  author={Baban Gain and Dibyanayan Bandyopadhyay and Subhabrata Mukherjee and Chandranath Adak and Asif Ekbal},
journal={arXiv preprint arXiv:2308.16075},
year={2023},
  url={https://arxiv.org/abs/2308.16075}
}

@inproceedings{2022-jian,
  author={Jian, Weichen and Hou, Hongxu and Wu, Nier and Sun, Shuo and Yang, ZongHeng and Wang, Yisong and Wang, Pengcong},
  booktitle={Proceedings of 2022 International Joint Conference on Neural Networks (IJCNN)}, 
  title="{Multimodal Neural Machine Translation for Mongolian to Chinese}", 
  year={2022},
  pages={1--8},
  doi={10.1109/IJCNN55064.2022.9892831},
  url={https://ieeexplore.ieee.org/document/9892831}}

@inproceedings{2024-yang,
   author={Yang, Yang and Ren, Qing-Dao-Er-Ji and He, Rui-Feng},
  booktitle={Proceedings of International Conference on Asian Language Processing (IALP)}, 
  title="{Multi-modal Sentiment Analysis of Mongolian Language based on Pre-trained Models and High-resolution Networks}", 
  year={2024},
  pages={291--296},
  doi={10.1109/IALP63756.2024.10661161}, 
  url={https://ieeexplore.ieee.org/document/10661161/}}

@inproceedings{2024-Alwajih,
       title = "Peacock: A Family of {A}rabic Multimodal Large Language Models and Benchmarks",
    author = "Alwajih, Fakhraddin  and
      Nagoudi, El Moatez Billah  and
      Bhatia, Gagan  and
      Mohamed, Abdelrahman  and
      Abdul-Mageed, Muhammad",
    
    booktitle = "Proceedings of the 62nd Annual Meeting of the Association for Computational Linguistics (ACL)",
    year = "2024",
    url = "https://aclanthology.org/2024.acl-long.689/",
    doi = "10.18653/v1/2024.acl-long.689",
    pages = "12753--12776"
}

@inproceedings{bhatia2024qalammultimodalllm,
       title = "Qalam: A Multimodal {LLM} for {A}rabic Optical Character and Handwriting Recognition",
    author = "Bhatia, Gagan  and
      Nagoudi, El Moatez Billah  and
      Alwajih, Fakhraddin  and
      Abdul-Mageed, Muhammad",
   
    booktitle = "Proceedings of the Second Arabic Natural Language Processing Conference (ArabicNLP)",
    year = "2024",
    url = "https://aclanthology.org/2024.arabicnlp-1.19/",
    doi = "10.18653/v1/2024.arabicnlp-1.19",
    pages = "210--224"
}

@article{2024-kovath,
author = {Kovath, Abhishek Gopinath and Nayyar, Anand and Sikha, O. K.},
title = "{Multimodal attention-driven visual question answering for Malayalam}",
year = {2024},
volume = {36},
number = {24},
issn = {0941-0643},
url = {https://link.springer.com/article/10.1007/s00521-024-09818-4},
doi = {10.1007/s00521-024-09818-4},
journal = {Neural Computing and Applications},
pages = {14691--14708},
numpages = {18},
}

@inproceedings{2024-lovenia,
 title = "{SEAC}rowd: A Multilingual Multimodal Data Hub and Benchmark Suite for {S}outheast {A}sian Languages",
    author = {Lovenia, Holy  and
      Mahendra, Rahmad  and
      Akbar, Salsabil Maulana  and
      Miranda, Lester James Validad  and
      Santoso, Jennifer  and
      Aco, Elyanah  and
      Fadhilah, Akhdan  and
      Mansurov, Jonibek  and
      Imperial, Joseph Marvin  and
      Kampman, Onno P.  and others},
   
    booktitle = "Proceedings of the 2024 Conference on Empirical Methods in Natural Language Processing (EMNLP)",
    year = "2024",
    url = "https://aclanthology.org/2024.emnlp-main.296/",
    doi = "10.18653/v1/2024.emnlp-main.296",
    pages = "5155--5203"
}

@inproceedings{2024-shen,
title = "The Language Barrier: Dissecting Safety Challenges of {LLM}s in Multilingual Contexts",
    author = "Shen, Lingfeng  and
      Tan, Weiting  and
      Chen, Sihao  and
      Chen, Yunmo  and
      Zhang, Jingyu  and
      Xu, Haoran  and
      Zheng, Boyuan  and
      Koehn, Philipp  and
      Khashabi, Daniel",
  
    booktitle = "Findings of the Association for Computational Linguistics (ACL)",
    year = "2024",
    url = "https://aclanthology.org/2024.findings-acl.156/",
    doi = "10.18653/v1/2024.findings-acl.156",
    pages = "2668--2680"
}

@inproceedings{2023-wchen,
  author={Chen, William and Yan, Brian and Shi, Jiatong and Peng, Yifan and Maiti, Soumi and Watanabe, Shinji},
  booktitle={Proceedings of IEEE International Conference on Acoustics, Speech and Signal Processing (ICASSP)}, 
  title="{Improving Massively Multilingual ASR with Auxiliary CTC Objectives}", 
  year={2023},
  pages={1--5},
  doi={10.1109/ICASSP49357.2023.10095326},
  url={https://ieeexplore.ieee.org/document/10095326}
}

@inproceedings{2024-yeo,
      title = "{Visual Speech Recognition for Languages with Limited Labeled Data Using Automatic Labels from Whisper}",
doi = {10.1109/ICASSP48485.2024.10446720}, 
      author={Yeo, Jeong and Kim, Minsu and Watanabe, Shinji and Ro, Yong},
  booktitle={Proceedings of IEEE International Conference on Acoustics, Speech and Signal Processing (ICASSP)}, 
      year={2024},
      pages={10471--10475},
      url={https://ieeexplore.ieee.org/document/10446720}
}

@article{2024-nortje,
      author = {Nortje, Leanne and Onea\c{t}\u{a}, Dan  and Kamper, Herman},
title = {Visually Grounded Few-Shot Word Learning in Low-Resource Settings},
year = {2024},
issue_date = {2024},
volume = {32},
issn = {2329-9290},
url = {https://ieeexplore.ieee.org/document/10508479/},
doi = {10.1109/TASLP.2024.3393772},
journal = {IEEE/ACM Transactions on Audio, Speech and Language Processing},
pages = {2544--2554},
numpages = {11}
}

@article{Pais2024RoMemesAM,
  title="{RoMemes: A multimodal meme corpus for the Romanian language}",
    author={P{\u{a}}i{\c{s}}, Vasile and Ni{\c{t}}{\u{a}}, Sara and Jerpelea, Alexandru-Iulius and Pan{\u{a}}, Luca and Curea, Eric},
  journal={arXiv preprint arXiv:2410.15497},
  year={2024},
  url={https://arxiv.org/abs/2410.15497}
}

@inproceedings{kponou-etal-2024-ffstc,
    title = "{FFSTC}: Fongbe to {F}rench Speech Translation Corpus",
    author = "Kponou, D. Fortun{\'e}  and
      Laleye, Fr{\'e}jus A.A.  and
      Ezin, Eug{\`e}ne Cokou",
    
    booktitle = "Proceedings of the 2024 Joint International Conference on Computational Linguistics, Language Resources and Evaluation (LREC-COLING)",
    year = "2024",
    url = "https://aclanthology.org/2024.lrec-main.638/",
    pages = "7270--7276"
}

@inproceedings{2024-rahman,
    title = "{B}inary{\_}{B}easts@{D}ravidian{L}ang{T}ech-{EACL} 2024: Multimodal Abusive Language Detection in {T}amil based on Integrated Approach of Machine Learning and Deep Learning Techniques",
    author = "Rahman, Md.  and
      Raihan, Abu  and
      Rahman, Tanzim  and
      Ahsan, Shawly  and
      Hossain, Jawad  and
      Das, Avishek  and
      Hoque, Mohammed Moshiul",
    booktitle = "Proceedings of the Fourth Workshop on Speech, Vision, and Language Technologies for Dravidian Languages (DravidianLangTech)",
    year = "2024",
    url = "https://aclanthology.org/2024.dravidianlangtech-1.35/",
    pages = "212--217"
}

@article{2024-doan,
      title="{Vintern-1B: An Efficient Multimodal Large Language Model for Vietnamese}", 
      author={Doan, Khang T. and Huynh, Bao G. and Hoang, Dung T. and Pham, Thuc D. and Pham, Nhat H. and Nguyen, Quan and Vo, Bang Q. and Hoang, Suong N.},
  journal={arXiv preprint arXiv:2408.12480},
      year={2024},
  url={https://arxiv.org/abs/2408.12480}, 
}

@inproceedings{radford2021learning,
   author       = {Alec Radford and
                  Jong Wook Kim and
                  Chris Hallacy and
                  Aditya Ramesh and
                  Gabriel Goh and
                  Sandhini Agarwal and
                  Girish Sastry and
                  Amanda Askell and
                  Pamela Mishkin and
                  Jack Clark and
                  Gretchen Krueger and
                  Ilya Sutskever},
  title        = {Learning Transferable Visual Models From Natural Language Supervision},
  booktitle    = {Proceedings of the 38th International Conference on Machine Learning (ICML)},
  volume       = {139},
  pages        = {8748--8763},
  year         = {2021},
  url          = {http://proceedings.mlr.press/v139/radford21a.html},
}

@inproceedings{devlin-etal-2019-bert,
    title = "{BERT}: Pre-training of Deep Bidirectional Transformers for Language Understanding",
    author = "Devlin, Jacob  and
      Chang, Ming-Wei  and
      Lee, Kenton  and
      Toutanova, Kristina",
    booktitle = "Proceedings of the 2019 Conference of the North {A}merican Chapter of the Association for Computational Linguistics: Human Language Technologies (NAACL-HLT)",
    year = "2019",
    url = "https://aclanthology.org/N19-1423/",
    doi = "10.18653/v1/N19-1423",
    pages = "4171--4186",
}

@inproceedings{2025-tayir,
author="Tayir, Turghun
and Li, Lin
and Maimaiti, Mieradilijiang
and Muhtar, Yusnur",
title="Low-Resource Machine Translation with Different Granularity Image Features",
booktitle="Proceedings of Chinese Conference on Pattern Recognition and Computer Vision (PRCV)",
year="2025",
pages="260--273",
isbn="978-981-97-8620-6",
url="https://link.springer.com/chapter/10.1007/978-981-97-8620-6_18"
}

@article{kirkpatrick2017overcoming,
  title={Overcoming catastrophic forgetting in neural networks},
  author={Kirkpatrick, James and Pascanu, Razvan and Rabinowitz, Neil and Veness, Joel and Desjardins, Guillaume and Rusu, Andrei A. and Milan, Kieran and Quan, John and Ramalho, Tiago and Grabska-Barwinska, Agnieszka and others},
  journal={Proceedings of the National Academy of Sciences},
  volume={114},
  number={13},
  pages={3521--3526},
  year={2017},
doi={10.1073/pnas.1611835114},
url={https://www.pnas.org/doi/full/10.1073/pnas.1611835114}
}

@article{2024-nortje-2,
      title={Improved Visually Prompted Keyword Localisation in Real Low-Resource Settings}, 
        author={Nortje, Leanne and Onea\c{t}\u{a}, Dan and Pirlogeanu, Gabriel and Kamper, Herman},
  journal={arXiv preprint arXiv:2409.06013},
      year={2024},
url={https://arxiv.org/abs/2409.06013},   
}

@article{krishna2017visual,
  author = {Krishna, Ranjay and Zhu, Yuke and Groth, Oliver and Johnson, Justin and Hata, Kenji and Kravitz, Joshua and Chen, Stephanie and Kalantidis, Yannis and Li, Li-Jia and Shamma, David A. and Bernstein, Michael S. and Fei-Fei, Li},
title = {Visual Genome: Connecting Language and Vision Using Crowdsourced Dense Image Annotations},
year = {2017},
issue_date = {May 2017},
volume = {123},
number = {1},
issn = {0920-5691},
url = {https://link.springer.com/article/10.1007/s11263-016-0981-7},
doi = {10.1007/s11263-016-0981-7},
journal = {International Journal of Computer Vision},
pages = {32--73},
numpages = {42},
}

@inproceedings{elliott2016multi30k,
      title = "{M}ulti30{K}: Multilingual {E}nglish-{G}erman Image Descriptions",
    author = "Elliott, Desmond  and
      Frank, Stella  and
      Sima{'}an, Khalil  and
      Specia, Lucia",
    booktitle = "Proceedings of the 5th Workshop on Vision and Language (VL'16)",
    year = "2016",
    url = "https://aclanthology.org/W16-3210/",
    doi = "10.18653/v1/W16-3210",
    pages = "70--74"
}

@article{2023-Yin,
      author = {Yin, Shukang and Fu, Chaoyou and Zhao, Sirui and Li, Ke and Sun, Xing and Xu, Tong and Chen, Enhong},
    title = {A survey on multimodal large language models},
    journal = {National Science Review},
    volume = {11},
    number = {12},
    pages = {nwae403},
    year = {2024},
    issn = {2095-5138},
    doi = {10.1093/nsr/nwae403},
    url = {https://academic.oup.com/nsr/article/11/12/nwae403/7896414},
}

@article{2023-Wang-survey,
  title={Large-scale multi-modal pre-trained models: A comprehensive survey},
  author = {Xiao Wang and Guangyao Chen and Guangwu Qian and Pengcheng Gao and Xiao-Yong Wei and Yaowei Wang and Yonghong Tian and Wen Gao},
  journal={Machine Intelligence Research},
    volume={20},
  pages={447--482},
  year={2023},
doi={10.1007/s11633-022-1410-8},
url={https://link.springer.com/article/10.1007/s11633-022-1410-8}
}

@article{zhang-2024,
  AUTHOR = {Zhang, Zhenwei and Zhang, Shengming and Ni, Dong and Wei, Zhaoguo and Yang, Kongjun and Jin, Shan and Huang, Gan and Liang, Zhen and Zhang, Li and Li, Linling and others},
TITLE = {Multimodal Sensing for Depression Risk Detection: Integrating Audio, Video, and Text Data},
JOURNAL = {Sensors},
VOLUME = {24},
YEAR = {2024},
NUMBER = {12},
pages = {3714},
URL = {https://www.mdpi.com/1424-8220/24/12/3714},
PubMedID = {38931497},
ISSN = {1424-8220},
DOI = {10.3390/s24123714}
}

@article{haputhanthri-2023,
  title="{Multi-modal Deep Learning Approach to Improve Sentence level Sinhala Sign Language Recognition}",
  author={Haputhanthri, H.H.S.N. and Tennakoon, H.M.N. and Wijesekara, M.A.S.M. and Pushpananda, B.H.R. and Thilini, H.N.D.},
  journal={International Journal on Advances in ICT for Emerging Regions},
  year={2023},
volume={16},
number={2},
  pages={21--30},
doi={10.4038/icter.v16i2.7264},
url={https://icter.sljol.info/articles/10.4038/icter.v16i2.7264}
}

@inproceedings{ristea2023cascaded,
    title={Cascaded cross-modal transformer for request and complaint detection},
  author={Ristea, Nicolae-Catalin and Ionescu, Radu Tudor},
  booktitle={Proceedings of the 31st ACM International Conference on Multimedia (ACMMM)},
  pages={9467--9471},
  year={2023},
  doi={10.1145/3581783.3612846},
  url={https://dl.acm.org/doi/10.1145/3581783.3612846}
}

@article{gu2024survey,
  title="{A Survey on LLM-as-a-Judge}",
  author={Jiawei Gu and Xuhui Jiang and Zhichao Shi and Hexiang Tan and Xuehao Zhai and Chengjin Xu and Wei Li and Yinghan Shen and Shengjie Ma and Honghao Liu and Yuanzhuo Wang and Jian Guo},
journal={arXiv preprint arXiv:2411.15594},
year={2024},
  url={https://arxiv.org/abs/2411.15594}
}

@article{xie2024large,
  title="{Large Multimodal Agents: A Survey}",
  author={Junlin Xie and Zhihong Chen and Ruifei Zhang and Xiang Wan and Guanbin Li},
journal={arXiv preprint arXiv:2402.15116},
year={2024},
  url={https://arxiv.org/abs/2402.15116}
}

@article{xu2024survey,
  title="{A Survey of Resource-efficient LLM and Multimodal Foundation Models}",
  author={Mengwei Xu and Wangsong Yin and Dongqi Cai and Rongjie Yi and Daliang Xu and Qipeng Wang and Bingyang Wu and Yihao Zhao and Chen Yang and Shihe Wang and others},
journal={arXiv preprint arXiv:2401.08092},
year={2024},
  url={https://arxiv.org/abs/2401.08092}
}

@article{solomon2023amharic,
  title={Amharic language image captions generation using hybridized attention-based deep neural networks},
  author={Solomon, Rodas and Abebe, Mesfin},
  journal={Applied Computational Intelligence and Soft Computing},
  volume={2023},
  number={1},
  pages={9397325},
  year={2023},
  doi={10.1155/2023/9397325},
  url={https://onlinelibrary.wiley.com/doi/10.1155/2023/9397325}
}

@article{rahul2023morphology,
  title="{Morphology \& word sense disambiguation embedded multimodal neural machine translation system between Sanskrit and Malayalam}",
  author={Rahul, C. and Arathi, T. and Panicker, Lakshmi S. and Gopikakumari, R.},
  journal={Biomedical Signal Processing and Control},
  volume={85},
  pages={105051},
  year={2023},
doi="10.1016/j.bspc.2023.105051",
url = {https://www.sciencedirect.com/science/article/pii/S1746809423004846},
}

@article{rahmon2024speech,
  title="{Speech Recognition Model in Yoruba Language}",
  author={Rahmon, Habeeb A. and Jimoh, Tope G. and Madaiyese, Fatimoh O.},
  journal={Smartify: Journal of Smart Education and Pedagogy},
  volume={1},
  number={1},
  pages={28--46},
  year={2024},
url="https://researchvision.us/index.php/smartify/article/view/5"
}

@inproceedings{meetei2021low,
  title="{Low resource multimodal neural machine translation of English-Hindi in news domain}",
  author={Meetei, Loitongbam Sanayai and Singh, Thoudam Doren and Bandyopadhyay, Sivaji},
  booktitle={Proceedings of the First Workshop on Multimodal Machine Translation for Low Resource Languages (MMTLRL)},
  pages={20--29},
  year={2021},
 url = "https://aclanthology.org/2021.mmtlrl-1.4/",
}

@inproceedings{wang2022cross,
  title={Cross-lingual cross-modal retrieval with noise-robust learning},
  author={Wang, Yabing and Dong, Jianfeng and Liang, Tianxiang and Zhang, Minsong and Cai, Rui and Wang, Xun},
  booktitle={Proceedings of the 30th ACM International Conference on Multimedia (ACMMM)},
  pages={422--433},
  year={2022},
  url = {https://dl.acm.org/doi/10.1145/3503161.3548003},
  doi = {10.1145/3503161.3548003} 
}

@inproceedings{nath2022image,
  title={Image caption generation for low-resource {A}ssamese language},
  author={Nath, Prachurya and Adhikary, Prottay Kumar and Dadure, Pankaj and Pakray, Partha and Manna, Riyanka and Bandyopadhyay, Sivaji},
  booktitle={Proceedings of the 34th Conference on Computational Linguistics and Speech Processing (ROCLING 2022)},
  pages={263--272},
  year={2022},
url="https://aclanthology.org/2022.rocling-1.33/"
}

@article{jiang2024multimodal,
author = {Lanlan Jiang and Jun Li and Jingwei Zhang and Yan Shen},
year = {2024},
title = "{Multimodal Seed Data Augmentation for Low-Resource Audio Latin Cuengh Language}",
journal = {Applied Sciences},
VOLUME = {14},
NUMBER = {20},
pages = {9533},
doi ={10.3390/app14209533},
url = {https://www.mdpi.com/2076-3417/14/20/9533}
}

@article{cheema2024adapting,
  title="{Adapting multilingual vision language transformers for low-resource Urdu optical character recognition (OCR)}",
  author={Cheema, Musa Dildar Ahmed and Shaiq, Mohammad Daniyal and Mirza, Farhaan and Kamal, Ali and Naeem, M. Asif},
  journal={PeerJ Computer Science},
  volume={10},
  pages={e1964},
  year={2024},
url={https://peerj.com/articles/cs-1964/}
}

@inproceedings{yang2022empirical,
  title="{An empirical study of GPT-3 for few-shot knowledge-based VQA}",
  author={Yang, Zhengyuan and Gan, Zhe and Wang, Jianfeng and Hu, Xiaowei and Lu, Yumao and Liu, Zicheng and Wang, Lijuan},
  booktitle={Proceedings of the AAAI Conference on Artificial Intelligence (AAAI)},
  volume={36},
  pages={3081--3089},
  year={2022},
  doi={10.1609/aaai.v36i3.20215},
  url={https://ojs.aaai.org/index.php/AAAI/article/view/20215}
}

@article{vetriselvi2024sentiment,
author = {Aruna Gladys, A. and Vetriselvi, V.},
title = {Sentiment analysis on a low-resource language dataset using multimodal representation learning and cross-lingual transfer learning},
year = {2024},
volume = {157},
number = {C},
issn = {1568-4946},
url = {https://www.sciencedirect.com/science/article/abs/pii/S1568494624003272},
doi = {10.1016/j.asoc.2024.111553},
journal = {Applied Soft Computing},
pages = {111553},
}

@article{das2022multi,
  author = {Das, Ringki and Singh, Thoudam Doren},
title = "{A multi-stage multimodal framework for sentiment analysis of Assamese in low resource setting}",
year = {2022},
volume = {204},
number = {C},
issn = {0957-4174},
url = {https://www.sciencedirect.com/science/article/abs/pii/S0957417422008879},
doi = {10.1016/j.eswa.2022.117575},
journal = {Expert Systems with Applications},
pages = {117575},
}

@article{deocampo2024lip,
	author = {Nikie Jo Deocampo and Mia Villarica and Albert Vinluan},
	title = "{A Lip-Reading Model for Tagalog Using Multimodal Deep Learning Approach}",
	journal = {International Journal of Computing Sciences Research},
	volume = {8},
	year = {2024},
	issn = {2546-115X},	
pages = {2796--2808},	
url = {https://stepacademic.net/ijcsr/article/view/511}
}

@article{mamyrbayev2020multimodal,
author = {Mamyrbayev, Orken Zh. and Alimhan, Keylan and Amirgaliyev, Beibut and Zhumazhanov, Bagashar and Mussayeva, Dinara and Gusmanova, Farida},
title = {Multimodal systems for speech recognition},
year = {2020},
issue_date = {2020},
volume = {18},
number = {3},
issn = {1470-949X},
url = {https://doi.org/10.1504/ijmc.2020.107097},
doi = {10.1504/ijmc.2020.107097},
journal = {International Journal of Mobile Communications},
pages = {314--326},
numpages = {12},
}

@inproceedings{kodali2025bytesizedllm,
  title="{byteSizedLLM@DravidianLangTech 2025: Abusive Tamil and Malayalam Text targeting Women on Social Media Using XLM-RoBERTa and Attention-BiLSTM}",
  author={Kodali, Rohith Gowtham and Manukonda, Durga Prasad and Pannakkaran, Maharajan},
  booktitle={Proceedings of the Fifth Workshop on Speech, Vision, and Language Technologies for Dravidian Languages (DravidianLangTech)},
  pages={80--85},
  year={2025},
url = "https://aclanthology.org/2025.dravidianlangtech-1.14/",
}

@article{faria2025sentimentformer,
AUTHOR = {Faria, Fatema Tuj Johora and Baniata, Laith H. and Baniata, Mohammad H. and Khair, Mohannad A. and Bani Ata, Ahmed Ibrahim and Bunterngchit, Chayut and Kang, Sangwoo},
TITLE = "{SentimentFormer: A Transformer-Based Multimodal Fusion Framework for Enhanced Sentiment Analysis of Memes in Under-Resourced Bangla Language}",
JOURNAL = {Electronics},
VOLUME = {14},
YEAR = {2025},
NUMBER = {4},
pages = {799},
URL = {https://www.mdpi.com/2079-9292/14/4/799},
DOI = {10.3390/electronics14040799}
}

@inproceedings{Ruder2021XTREMERTM,
title="{{XTREME}-{R}: Towards More Challenging and Nuanced Multilingual Evaluation}",
  author = "Ruder, Sebastian  and
      Constant, Noah  and
      Botha, Jan  and
      Siddhant, Aditya  and
      Firat, Orhan  and
      Fu, Jinlan  and
      Liu, Pengfei  and
      Hu, Junjie  and
      Garrette, Dan  and
      Neubig, Graham  and
      Johnson, Melvin",
    booktitle = "Proceedings of the 2021 Conference on Empirical Methods in Natural Language Processing (EMNLP)",
    year = "2021",
    url = "https://aclanthology.org/2021.emnlp-main.802/",
    doi = "10.18653/v1/2021.emnlp-main.802",
    pages = "10215--10245",
}

@article{Paullada2020DataAI,
title = {Data and its (dis)contents: A survey of dataset development and use in machine learning research},
journal = {Patterns},
volume = {2},
number = {11},
pages = {100336},
year = {2021},
issn = {2666-3899},
doi = {https://doi.org/10.1016/j.patter.2021.100336},
url = {https://www.sciencedirect.com/science/article/pii/S2666389921001847},
author = {Amandalynne Paullada and Inioluwa Deborah Raji and Emily M. Bender and Emily Denton and Alex Hanna},
}

@inproceedings{joshi2020state,
   title={The State and Fate of Linguistic Diversity and Inclusion in the {NLP} World},
  author={Pratik M. Joshi and Sebastin Santy and Amarjit Budhiraja and Kalika Bali and Monojit Choudhury},
  booktitle={Proceedings of the 58th Annual Meeting of the Association for Computational Linguistics (ACL)},
  year={2020},
  pages={6282--6293},
  url={https://aclanthology.org/2020.acl-main.560/},
  doi={10.18653/v1/2020.acl-main.560}
}

@article{Zhao2023ASO,
  title="{A Survey of Large Language Models}",
  author={Wayne Xin Zhao and Kun Zhou and Junyi Li and Tianyi Tang and Xiaolei Wang and Yupeng Hou and Yingqian Min and Beichen Zhang and Junjie Zhang and Zican Dong and others},
journal={arXiv preprint arXiv:2303.18223},
year={2023},
  url={https://arxiv.org/abs/2303.18223}
}

@article{Zhu2023ExtrapolatingLL,
  title="{Extrapolating Large Language Models to Non-English by Aligning Languages}",
  author={Wenhao Zhu and Yunzhe Lv and Qingxiu Dong and Fei Yuan and Jingjing Xu and Shujian Huang and Lingpeng Kong and Jiajun Chen and Lei Li},
journal={arXiv preprint arXiv:2308.04948},
year={2023},
  url={https://arxiv.org/abs/2308.04948}
}

@article{Gan2022VisionLanguagePB,
author = {Gan, Zhe and Li, Linjie and Li, Chunyuan and Wang, Lijuan and Liu, Zicheng and Gao, Jianfeng},
title = {Vision-Language Pre-Training: Basics, Recent Advances, and Future Trends},
year = {2022},
issue_date = {Dec 2022},
volume = {14},
number = {3–4},
issn = {1572-2740},
url = {https://www.nowpublishers.com/article/Details/CGV-105},
doi = {10.1561/0600000105},
journal = {Foundations and Trends in Computer Graphics and Vision},
pages = {163--352},
numpages = {204}
}

@article{shukor2025scaling_data,
  title={Scaling Laws for Optimal Data Mixtures},
  author={Shukor, Mustafa and Bethune, Louis and Busbridge, Dan and Grangier, David and Fini, Enrico and El-Nouby, Alaaeldin and Ablin, Pierre},
  journal={arXiv preprint arXiv:2507.09404},
  year={2025},
url={https://arxiv.org/abs/2507.09404}
}

@inproceedings{Mixture_of_LoRA,
  author       = {Ying Shen and
                  Zhiyang Xu and
                  Qifan Wang and
                  Yu Cheng and
                  Wenpeng Yin and
                  Lifu Huang},
  title        = "{Multimodal Instruction Tuning with Conditional Mixture of LoRA}",
  booktitle    = {Proceedings of the 62nd Annual Meeting of the Association for Computational Linguistics (ACL)},
  pages        = {637--648},
  year         = {2024},
  url          = {https://aclanthology.org/2024.acl-long.38/},
  doi          = {10.18653/V1/2024.ACL-LONG.38},
}

@article{lamma,
  author       = {Abhimanyu Dubey and
                  Abhinav Jauhri and
                  Abhinav Pandey and
                  Abhishek Kadian and
                  Ahmad Al{-}Dahle and
                  Aiesha Letman and
                  Akhil Mathur and
                  Alan Schelten and
                  Amy Yang and
                  Angela Fan and
                 others},
  title        = {The {L}lama 3 Herd of Models},
journal={arXiv preprint arXiv:2407.21783},
  year         = {2024},
  url          = {https://arxiv.org/abs/2407.21783},
  doi          = {10.48550/ARXIV.2407.21783},
}

@article{DeepSeek,
  author       = {DeepSeek{-}AI and
                  Aixin Liu and
                  Bei Feng and
                  Bing Xue and
                  Bingxuan Wang and
                  Bochao Wu and
                  Chengda Lu and
                  Chenggang Zhao and
                  Chengqi Deng and
                  Chenyu Zhang and
                  others},
  title        = {DeepSeek-V3 Technical Report},
  journal={arXiv preprint arXiv:2412.19437},
  year         = {2024},
  url          = {https://arxiv.org/abs/2412.19437},
  doi          = {10.48550/ARXIV.2412.19437},
}

@misc{anthropic2025claude,
  author       = {Anthropic},
  title        = {Claude {O}pus 4 \& {C}laude {S}onnet 4 System Card},
  year         = {2025},
 
  url          = {https://www-cdn.anthropic.com/07b2a3f9902ee19fe39a36ca638e5ae987bc64dd.pdf},
  note         = {Accessed: December 2025}
}

@article{DBLP:journals/corr/abs-2407-21075,
  author       = {Tom Gunter and
                  Zirui Wang and
                  Chong Wang and
                  Ruoming Pang and
                  Andy Narayanan and
                  Aonan Zhang and
                  Bowen Zhang and
                  Chen Chen and
                  Chung{-}Cheng Chiu and
                  David Qiu and
                  others},
  title        = "{Apple Intelligence Foundation Language Models}",
  journal={arXiv preprint arXiv:2407.21075},
  year         = {2024},
  url          = {https://arxiv.org/abs/2407.21075},
  doi          = {10.48550/ARXIV.2407.21075},
}

@article{MMaDA,
  author       = {Ling Yang and
                  Ye Tian and
                  Bowen Li and
                  Xinchen Zhang and
                  Ke Shen and
                  Yunhai Tong and
                  Mengdi Wang},
  title        = "{MMaDA: Multimodal Large Diffusion Language Models}",
  journal={arXiv preprint arXiv:2505.15809},
  year         = {2025},
  url          = {https://arxiv.org/abs/2505.15809},
  doi          = {10.48550/ARXIV.2505.15809},
  }

@inproceedings{Art_Large_Vision,
  author       = {Zongxia Li and
                  Xiyang Wu and
                  Hongyang Du and
                  Fuxiao Liu and
                  Huy Nghiem and
                  Guangyao Shi},
  title        = "{A Survey of State of the Art Large Vision Language Models: Benchmark Evaluations and Challenges}",
  booktitle    = {Proceedings of {IEEE/CVF} Conference on Computer Vision and Pattern Recognition Workshops (CVPRW)},
  pages        = {1587--1606},
  year         = {2025},
  url          = {https://openaccess.thecvf.com/content/CVPR2025W/TMM-OpenWorld/html/Li\_A\_Survey\_of\_State\_of\_the\_Art\_Large\_Vision\_Language\_CVPRW\_2025\_paper.html},
}

@article{DBLP:journals/corr/abs-2503-07137,
  author       = {Siyuan Mu and
                  Sen Lin},
  title        = "{A Comprehensive Survey of Mixture-of-Experts: Algorithms, Theory, and Applications}",
   journal={arXiv preprint arXiv:2503.07137},
  year         = {2025},
  url          = {https://doi.org/10.48550/arXiv.2503.07137},
  doi          = {10.48550/ARXIV.2503.07137},
}

@article{carroll2020care,
  title={The {CARE} Principles for Indigenous Data Governance},
  author={Stephanie Russo Carroll and Ibrahim Garba and Oscar L. Figueroa-Rodr{\'i}guez and Jarita C. Holbrook and Raymond Lovett and Simeon Materechera and Mark A. Parsons and Kay Raseroka and Desi Rodriguez-Lonebear and Robyn Rowe and others},
  journal={Data Science Journal},
  year={2020},
  volume={19},
  pages={43},
  url={https://datascience.codata.org/articles/dsj-2020-043},
  doi={10.5334/dsj-2020-043}
}

@inproceedings{wiechetek2024ethical,
  title={The Ethical Question – Use of Indigenous Corpora for Large Language Models},
  author={Linda Wiechetek and Flammie A. Pirinen and B{\o}rre Gaup and Trond Trosterud and Maja Kappfjell and Sjur N{\o}rsteb{\o} Moshagen},
  booktitle={Proceedings of the 2024 Joint International Conference on Computational Linguistics, Language Resources and Evaluation (LREC-COLING)},
  year={2024},
  pages={15922--15931},
  url={https://aclanthology.org/2024.lrec-main.1383/}
}

@article{navigli2023biases,
  title={Biases in Large Language Models: Origins, Inventory, and Discussion},
  author={Roberto Navigli and Simone Conia and Bj{\"o}rn Ross},
  journal={ACM Journal of Data and Information Quality},
  year={2023},
  volume={15},
  pages={1--21},
  url={https://dl.acm.org/doi/10.1145/3597307},
  doi={10.1145/3597307}
}

@article{wu2025inequalities,
  title={Uncovering inequalities in new knowledge learning by large language models across different languages},
  author={Chenglong Wang and Haoyu Tang and Xiyuan Yang and Yueqi Xie and Jina Suh and Sunayana Sitaram and Junming Huang and Yueqi Xie and Zhaoya Gong and Xing Xie and Fangzhao Wu},
  journal={arXiv preprint arXiv:2503.04064},
  year={2025},
  url={https://arxiv.org/abs/2503.04064}
}

@article{dotan2024invisible,
  title={Invisible Languages of the {LLM} Universe},
  author={Saurabh Khanna and Xinxu Li},
  journal={arXiv preprint arXiv:2510.11557},
  year={2025},
  url={https://arxiv.org/abs/2510.11557}
}

@article{chen2025breaking,
  title={Breaking Physical and Linguistic Borders: Multilingual Federated Prompt Tuning for Low-Resource Languages},
  author={Wanru Zhao and Yihong Chen and Royson Lee and Xinchi Qiu and Yan Gao and Hongxiang Fan and Nicholas Donald Lane},
  journal={arXiv preprint arXiv:2507.03003},
  year={2025},
  doi={10.48550/arXiv.2507.03003},
  url={https://arxiv.org/abs/2507.03003}
}

@article{yang2025dpfpl,
  title={Privacy-Preserving Personalized Federated Prompt Learning for Multimodal Large Language Models},
  author={Linh Tran and Wei Sun and Stacy Patterson and Ana Milanova},
  journal={arXiv preprint arXiv:2501.13904},
  year={2025},
  doi={10.48550/arXiv.2501.13904},
  url={https://arxiv.org/abs/2501.13904}
}

@inproceedings{lin2022fednlp,
  title={{FedNLP}: Benchmarking Federated Learning Methods for Natural Language Processing Tasks},
  author={Lin, Bill Yuchen  and
      He, Chaoyang  and
      Ze, Zihang  and
      Wang, Hulin  and
      Hua, Yufen  and
      Dupuy, Christophe  and
      Gupta, Rahul  and
      Soltanolkotabi, Mahdi  and
      Ren, Xiang  and
      Avestimehr, Salman},
  booktitle={Findings of the Association for Computational Linguistics (NAACL)},
  year={2022},
  pages={157-175},
  doi={10.18653/v1/2022.findings-naacl.13},
  url={https://aclanthology.org/2022.findings-naacl.13/}
}

@inproceedings{singh2024globalmmu,
  title={{Global MMLU}: Understanding and Addressing Cultural and Linguistic Biases in Multilingual Evaluation},
  author={Shivalika Singh and Angelika Romanou and Cl{\'e}mentine Fourrier and David Ifeoluwa Adelani and Jian Gang Ngui and Daniel Vila-Suero and Peerat Limkonchotiwat and Kelly Marchisio and Wei Qi Leong and Yosephine Susanto and others},
  booktitle={Proceedings of the 63rd Annual Meeting of the Association for Computational Linguistics (ACL)},
  pages={18761--18799},
  year={2025},
  doi={10.18653/v1/2025.acl-long.919},
  url={https://aclanthology.org/2025.acl-long.919/}
}

@inproceedings{bird2022decolonising,
  title={Local Languages, Third Spaces, and other High-Resource Scenarios},
  author={Steven Bird},
  booktitle={Proceedings of the 60th Annual Meeting of the Association for Computational Linguistics (ACL)},
  year={2022},
  pages={7817–7829},
  url={https://aclanthology.org/2022.acl-long.539/},
  doi={10.18653/v1/2022.acl-long.539}
}

@inproceedings{rungta2022geographic,
  title={Geographic Citation Gaps in {NLP} Research},
  author={Mukund Rungta and Janvijay Singh and Saif M. Mohammad and Diyi Yang},
  booktitle={Proceedings of the 2022 Conference on Empirical Methods in Natural Language Processing (EMNLP)},
  pages={1371--1383},
  year={2022},
  url={https://aclanthology.org/2022.emnlp-main.89/},
  doi={10.18653/v1/2022.emnlp-main.89}
}

@inproceedings{ranathunga2022languages,
  title={Some Languages are More Equal than Others: Probing Deeper into the Linguistic Disparity in the {NLP} World},
  author={Surangika Ranathunga and Nisansa de Silva},
  booktitle={Proceedings of the 2nd Conference of the Asia-Pacific Chapter of the Association for Computational Linguistics and the 12th International Joint Conference on Natural Language Processing (AACL-IJCNLP)},
  pages={823--848},
  year={2022},
url = "https://aclanthology.org/2022.aacl-main.62/",
    doi = "10.18653/v1/2022.aacl-main.62",
}

@misc{caines2019geographic,
  title={The Geographic Diversity of {NLP} Conferences},
  author={Caines, Andrew and Rei, Marek},
  year={2019},
  howpublished={\url{https://www.marekrei.com/blog/geographic-diversity-of-nlp-conferences/}},
  note={Accessed: December 2025}
}

@article{habash2022panoramic,
  title={A panoramic survey of natural language processing in the {A}rab world},
  author={Kareem Darwish and Nizar Habash and Mourad Abbas and Hend Suliman Al-Khalifa and Huseein T. Al-Natsheh and Samhaa R. El-Beltagy and Houda Bouamor and Karim Bouzoubaa and Violetta Cavalli-Sforza and Wassim El-Hajj and Mustafa Jarrar and Hamdy Mubarak},
  journal={Communications of the ACM},
  year={2020},
  volume={64},
  pages={72--81},
  url={https://dl.acm.org/doi/10.1145/3447735},
  doi={10.1145/3447735}
}

@misc{clarin2023ukrainian,
  title={{CLARIN} Knowledge Centre for {Ukrainian} {NLP} and Corpora ({UkrNLP-Corpora})},
  author={Olha Kanishcheva},
  year={2025},
  howpublished={\url{https://www.clarin.eu/blog/introduction-clarin-knowledge-centre-ukrainian-nlp-and-corpora-ukrnlp-corpora}},
  note={Accessed: December 2025}
}

@proceedings{unlp2025proceedings,
    title = "Proceedings of the Fourth Ukrainian Natural Language Processing Workshop (UNLP)",
    editor = "Romanyshyn, Mariana",
    year = "2025",
    publisher = "Association for Computational Linguistics",
    url = "https://aclanthology.org/2025.unlp-1.0/",
    doi = "10.18653/v1/2025.unlp-1.0",
    ISBN = "979-8-89176-269-5"
}

@inproceedings{akhynko2025hidden,
  title="{Hidden Persuasion: Detecting Manipulative Narratives on Social Media During the 2022 {R}ussian Invasion of {U}kraine}",
  author={Kateryna Akhynko and Oleksandr Kosovan and Mykola Trokhymovych},
  booktitle = "Proceedings of the Fourth Ukrainian Natural Language Processing Workshop (UNLP)",
  year={2025},
  pages={194--202},
  url={https://aclanthology.org/2025.unlp-1.19/},
  doi={10.18653/v1/2025.unlp-1.19}
}

@misc{secondtalent2025chinese,
  title={Top 50+ {Chinese} {AI} Investment Statistics [2025]},
  author={Matt Li},
  year={2025},
  howpublished={\url{https://www.secondtalent.com/resources/chinese-ai-investment-statistics/}},
  note={Accessed: December 2025}
}

@article{science2025deepseek,
  title={{Chinese} firm's faster, cheaper {AI} language model makes a splash},
  author={Normile, Dennis},
  journal={Science},
  year={2025},
volume={387},
number={6731},
pages={238--238},
  url={https://www.science.org/content/article/chinese-firm-s-faster-cheaper-ai-language-model-makes-splash}
}

\end{document}